\documentclass[twoside,11pt]{article}

\usepackage{blindtext}

\usepackage[preprint]{dmlr2e}

\usepackage[utf8]{inputenc} 
\usepackage[T1]{fontenc}    
\usepackage{hyperref}       
\usepackage{url}            
\usepackage{booktabs}       
\usepackage{amsfonts}       
\usepackage{nicefrac}       
\usepackage{microtype}      
\usepackage{xcolor}         
\usepackage{graphicx}
\usepackage{caption}
\usepackage{amsfonts}
\usepackage{wrapfig}
\usepackage{tabularx}
\usepackage{amsmath}
\usepackage{enumitem}
\usepackage{multirow}
\usepackage{multicol}

\newif\ifshowcomments
\showcommentstrue 
\showcommentsfalse 
\ifshowcomments
\newcommand {\yaoqing}[1]{{\color{orange}\sf{[Yaoqing: #1]}}}
\newcommand {\konstantin}[1]{{\color{red}\sf{[Konstantin: #1]}}}
\newcommand {\yefan}[1]{{\color{purple}\sf{[Yefan: #1]}}}
\definecolor{gold}{HTML}{DAA520}
\newcommand {\leo}[1]{{\color{gold}\sf{[Léo: #1]}}}
\else
\newcommand {\yaoqing}[1]{}
\newcommand {\konstantin}[1]{}
\newcommand {\yefan}[1]{}
\newcommand{\leo}[1]{}
\fi

\usepackage{amsmath,amsfonts,bm}

\def\eqref#1{equation~\ref{#1}}

\def\1{\bm{1}}

\def\rmH{{\mathbf{H}}}

\def\rmK{{\mathbf{K}}}
\def\rmL{{\mathbf{L}}}

\def\rmX{{\mathbf{X}}}
\def\rmY{{\mathbf{Y}}}

\def\vtheta{{\bm{\theta}}}

\DeclareMathAlphabet{\mathsfit}{\encodingdefault}{\sfdefault}{m}{sl}
\SetMathAlphabet{\mathsfit}{bold}{\encodingdefault}{\sfdefault}{bx}{n}

\usepackage{lastpage}
\dmlrheading{XX}{2025}{1-\pageref{LastPage}}{03/25; Revised 09/25}{09/25}{21-0000}{K. Schürholt, L. Meynent, Y. Zhou, H. Lu, Y. Yang and D. Borth} 
\def\openreview{\url{https://openreview.net/forum?id=zJRWvNpdIr}} 

\ShortHeadings{Phase Transitions Model Zoo}{Schürholt, Meynent, Zhou, Lu, Yang and Borth} 
\firstpageno{1}

\begin{document}

\title{A Model Zoo on Phase Transitions in Neural Networks}

\author{\name Konstantin Schürholt \email konstantin.schuerholt@unisg.ch \\
       \addr Department of Computer Science\\
       University of St. Gallen, Switzerland
       \AND
       \name Léo Meynent \email leo.meynent@unisg.ch \\
       \addr Department of Computer Science\\
       University of St. Gallen, Switzerland
       \AND
       \name Yefan Zhou \email yefan.zhou.gr@dartmouth.edu \\
       \addr Department of Computer Science\\
       Dartmouth College, USA
       \AND
       \name Haiquan Lu \email haiquanlu@u.nus.edu \\
       \addr Department of Electrical and Computer Engineering\\
       National University of Singapore, Singapore
       \AND
       \name Yaoqing Yang \email yaoqing.yang@dartmouth.edu \\
       \addr Department of Computer Science\\
       Dartmouth College, USA
       \AND
       \name Damian Borth \email damian.borth@unisg.ch \\
       \addr Department of Computer Science\\
       University of St. Gallen, Switzerland
       }
       
\editor{Hongyang R. Zhang}

\maketitle

\begin{abstract}
Using the weights of trained Neural Network (NN) models as data modality has recently gained traction as a research field --- dubbed \textit{Weight Space Learning} (WSL).
Multiple recent works propose WSL methods to analyze models, evaluate methods, or synthesize weights.
Weight space learning methods require populations of trained models as datasets for development and evaluation.
However, existing collections of models --- called `model zoos' --- are unstructured or follow a rudimentary definition of diversity. 
In parallel, work rooted in statistical physics has identified phases and phase transitions in NN models. Models are homogeneous within the same phase but qualitatively differ from one phase to another.
We combine the idea of `model zoos' with phase information to create a controlled notion of diversity in populations.
We introduce 12 large-scale zoos that systematically cover known phases and vary over model architecture, size, and datasets. These datasets cover different modalities, such as computer vision, natural language processing, and scientific ML.
For every model, we compute loss landscape metrics and validate full coverage of the phases. 
With this dataset, we provide the community with a resource with a wide range of potential applications for WSL and beyond.
Evidence suggests the loss landscape phase plays a role in applications such as model training, analysis, or sparsification.
We demonstrate this in an exploratory study of the downstream methods like transfer learning or model weights averaging.
\looseness-1
\end{abstract}

\begin{keywords}
  Model Zoo, Weight Space Learning, Neural Networks, Phase Transition
\end{keywords}

\section{Introduction}

\begin{figure*}[t!]
\centering
\includegraphics[width=0.95\linewidth]{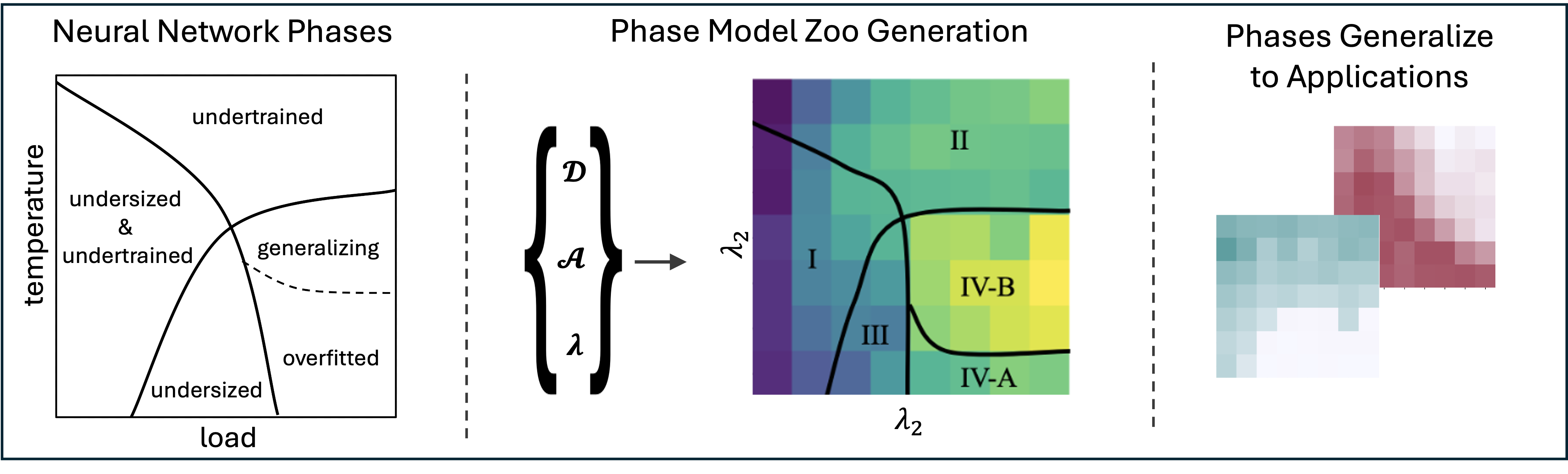}
\caption{
Overview of the approach. \textbf{Left:} Phases of Neural Networks with qualitatively different properties have been identified~\citep{martinRethinkingGeneralizationRequires2019,yangTaxonomizingLocalGlobal2021}. \textbf{Middle:} We propose to use phases as a quantifiable diversity metric and generate datasets of trained models by varying their datasets $\mathcal{D}$, architectures $\mathcal{A}$ and training parameters $\lambda$ s.t. they cover all known phases across different domains. \textbf{Right:} Phases manifest beyond evaluation metrics in applications of pre-trained models, e.g., in weight averaging. Our Phase Model Zoos allow systematic evaluation of methods that use pre-trained models.
}
\label{fig:approach_overview}
\end{figure*}

Training NNs has become standard practice on many tasks.
However, there are aspects of model training as well as the trained models that are still not fully understood. 
How to parameterize training and what model size to use remains a question for researchers and practitioners. 
With models growing in size, the computing power required to train them also increases exponentially~\citep{openai-compute}. It is therefore becoming more and more common to reuse pre-trained models from open model repositories like Hugging Face~\citep{huggingface} and fine-tune them for more specific tasks. 
However, the choice of a specific pre-trained model for fine-tuning or transfer learning strategies is non-trivial.

\paragraph{Weight Space Learning and Model Analysis.}
There is growing research interest in addressing these gaps by analyzing collections of trained models. Using neural networks as data modality -- dubbed \textit{weight space learning} (WSL) has become its own research field\footnote{ICLR 2025 hosts the first workshop on Neural Networks as Data Modality \url{https://weight-space-learning.github.io/}}.
Within the field of WSL, populations of models are used in a variety of domains: Applications include predicting how well models generalize~\citep{jiang2020neurips}, or detecting poisoned models~\citep{trojai}. Other methods use WSL torwards Neural Architecture Search to identify the relation between weights and model performance~\citep{martinPredictingTrendsQuality2021,zhou2024MD,liu2024model}. 
Different work re-uses existing trained weights to extend conventional transfer learning strategies~\citep{yosinskiHowTransferableAre2014, mensinkFactorsInfluenceTransfer2021} by combining multiple models, via a combination of their outputs~\citep{polikar2012ensemble, ganaie2022ensemble, mohammed2023comprehensive}, knowledge distillation~\citep{liuKnowledgeFlowImprove2019,shuZooTuningAdaptiveTransfer2021}, or by directly combining their weights~\citep{wortsmanModelSoupsAveraging2022, wortsmanRobustFinetuningZeroshot2022, mitchellFastModelEditing2022, ainsworthGitReBasinMerging2022, ilharcoEditingModelsTask2022, rameModelRatatouilleRecycling2023, rameWARMBenefitsWeight2024, navonEquivariantDeepWeight2024}, to build combined models with improved in- and out-of-distribution performance.\looseness-1

There is also a recent interest in using model populations to infer model properties from weights or activations~\citep{yakTaskArchitectureIndependentGeneralization2018, jiang2020neurips, eilertsenClassifyingClassifierDissecting2020, martinPredictingTrendsQuality2021}. Closely related, there is research on learning representations of NNs, for either model analysis~\citep{schurholtSelfSupervisedRepresentationLearning2021,ashkenaziNeRNLearningNeural2022,navonEquivariantArchitecturesLearning2023,zhouPermutationEquivariantNeural2023,deluigiDeepLearningImplicit2023,limGraphMetanetworksProcessing2023,andreisSetbasedNeuralNetwork2023,herrmannLearningUsefulRepresentations2024,navonEquivariantDeepWeight2024} or model generation~\citep{haHyperNetworks2017,zhangGraphHyperNetworksNeural2019,knyazevParameterPredictionUnseen2021,schurholtHyperRepresentationsGenerativeModels2022,peeblesLearningLearnGenerative2022,knyazevCanWeScale2023,zhangNeuralNetworksAre2023, shamsianImprovedGeneralizationWeight2024, schurholtScalableVersatileWeight2024, putterman2024learninglorasglequivariantprocessing, wang2024recurrent, soro2025diffusionbased}.\looseness-1

\paragraph{Phase Transitions of NNs.}
Another line of research investigates phase transitions in NNs and connects them to loss landscape metrics~\citep{martinRethinkingGeneralizationRequires2019,yangTaxonomizingLocalGlobal2021}. 
Motivated by statistical physics, they identify two main types of hyperparameters in NN training: the noisiness of the training process, dubbed temperature, and the amount of data relative to the size of the model, dubbed relative load. 
Using that notion, they identify distinct phases with qualitatively different model properties on the temperature-load landscape~\citep{yangTaxonomizingLocalGlobal2021}. 
Phase transitions have been studied extensively in the machine learning literature~\citep{schwarze1992generalization,seung1992statistical,martinRethinkingGeneralizationRequires2019,martinTraditionalHeavyTailedSelf2019}. While phase transitions are typically studied theoretically under certain limits, e.g., an infinitely wide NN~\citep{lewkowycz2020large}, we observe qualitatively similar properties in models with a practical size.
Within phases, models are relatively homogeneous, with abrupt changes from one phase to the next. Along the load dimension, one example of a phase transition is \emph{double descent}~\citep{nakkiranDeepDoubleDescent2019}. Neural scaling laws describe the load and temperature relation, e.g., as power laws to train models to specific phases~\citep{hestnessDeepLearningScaling2017,rosenfeldConstructivePredictionGeneralization2020,gordonDataParameterScaling2021,hoffmannTrainingComputeOptimalLarge2022,sorscherNeuralScalingLaws2022}. \looseness-1

\paragraph{Exisiting populations of Neural Networks.}
The field of \textit{weight space learning} uses models as data, and thus requires datasets of trained models.
Most freely accessible models are part of large public repositories like Hugging Face~\citep{huggingface} or the PyTorch model hub~\citep{pytorchcv2018}, opening up new opportunities such as transfer learning~\citep{prottasha2022transfer, lyu2021zero} and benchmarking~\citep{chiang2024chatbot, zheng2024judging}. Within those collections, however, models are of varying quality and mostly unstructured. The hyperparameters, data, and training strategy are often documented rudimentary, which makes them unsuitable for research on model populations. Researchers studying model populations sometimes publish the models they used for their work --- but the design of these populations and the available information are usually catered to their specific goal.

To fill that gap, structured populations have been published as model zoo datasets, in domains as varied as computer vision~\citep{unterthinerPredictingNeuralNetwork2020,eilertsenClassifyingClassifierDissecting2020,schurholtModelZoosDataset2022, falk2025model}, adversarial robustness assessment~\citep{croce2020robustbench}, bioimaging~\citep{ouyang2022bioimage} or Earth observation~\citep{honeggerEurosatModelZoo2023, honeggerSparsifiedModelZoo2023}. These populations, however, contain only models trained on one domain, and consider the diversity of the model in their populations only in terms of their generating factors. They report the distribution in model performance and model similarity, but it remains unclear whether that indicates conceptually different models, or whether they are variations of similar representations.

\newpage

\paragraph{Our WSL dataset covers multiple phases.}
To address that gap in datasets for \textit{weight space learning}, we combine the concept of model zoos with loss landscape and phase information~\citep{yangTaxonomizingLocalGlobal2021} and augment the model zoo blueprint with a new notion of diversity. Instead of aiming for diversity in model performance, which may or may not contain qualitatively different models, we create the model zoos with different architectures, model sizes, tasks and domains such that all the phases on the loss landscape are covered. This allows us to quantitatively show that the distinct, qualitatively different phases of models exist in our model zoos and generalize across architectures and domains.

Going beyond the models at regular training stage, our work shows the existence of phases in downstream methods such as fine-tuning, transfer learning, pruning, ensembling, etc. Such methods are usually evaluated on one or a few models, without explicit consideration for their phase diversity. Without that context, it is hard to attribute performance gains or the lack thereof to the method or the phase of the pre-trained model.
Model zoos that cover all phases could help systematically evaluate methods that rely on pre-trained models and identify where they work. This, in turn, may not only provide performance and robustness information but also a good signal to guide further research in this area. \looseness-1

In summary, with this dataset, we make the following contributions:
\vspace{-4pt}
\begin{itemize}[left=0pt]
    \setlength\itemsep{1pt}    
    \item We propose a new notion of diversity in model zoos by covering different model phases that relate to qualitatively different model properties using loss landscape taxonomy.\looseness-1
    \item We systematically create 12 model zoos that include all phases. The zoos cover computer vision, natural language and scientific machine learning (SciML) models, contain different architectures of different sizes and are trained on various datasets. It contains a total of $\sim$2.5k unique neural network models between 11K and 900M parameters and more than 60k checkpoints.
    \item We annotate the models with performance and loss landscape metrics, and include checkpoints for multiple epochs. We quantitatively validate that our zoos are diverse and cover known phases.
    \item We discuss the benefit of our dataset for the ML community: as quantifiable diverse model dataset for weight space learning, and contributing to better and more systematic evaluation of methods that use pre-trained models. 
    \item We make our dataset available at \url{https://phasetransitions.modelzoos.cc}
\end{itemize}

In the following, we first introduce metrics to taxonomize phases in loss landscapes (Sec. \ref{sec:loss-landscape}).
Subsequenlty, we detail the generation of the model zoos (Sec. \ref{sec:model_zoo_gen}).
We then evaluate that phases exist in our model zoos, generalize across different zoos, across domains, architectures, tasks and model sizes (Sec. \ref{sec:phases_empirics}).
Finally, we discuss concrete applications in weight space learning for our model zoos (Sec. \ref{sec:app_methods}).

\section{Loss Landscape Taxonomy}
\label{sec:loss-landscape}
\paragraph{Phases in Neural Networks Loss Landscapes.}
The motivation for introducing phases and phase transitions in NN loss landscapes is rooted in statistical mechanics, where such phenomena explain qualitative changes in system behavior~\citep{martinRethinkingGeneralizationRequires2019}.
Phases represent distinct regions in the parameter space where the system's properties are homogeneous or change smoothly, while phase transitions mark abrupt changes in these properties. 
In NNs, phases manifest in terms of generalization performance. A prominent example for such a phase transition is the double descent pheonmenon~\citep{nakkiranDeepDoubleDescent2019}, a phase transition along the axis of model size~\citep{belkinReconcilingModernMachinelearning2019, liao2020random, derezinski2020exact}. 
Similar empirical observations have been made recently on the \textit{emergent} abilities of large language models~\citep{wei2022emergent}, in which non-smooth transitions can occur when some training hyperparameters (such as the model size) are modified. However, it is not conclusive whether these emergent abilities are indeed sharp phase transitions or merely due to specific ways of experimental measurements~\citep{schaeffer2024emergent}.
These phases and transitions are expected due to the complex, high-dimensional nature of NN optimization, where varying control parameters like data noise and training iterations can lead to qualitatively different properties, akin to physical systems undergoing phase changes. 
Motivated by statistical physics,~\citet{martinRethinkingGeneralizationRequires2019} identify two main types of hyperparameters in NN training: the noisiness of the training process, dubbed temperature, and the amount of data relative to the size of the model, dubbed relative load. 
Using that notion, distinct phases with qualitatively different model properties on the temperature-load landscape can be identified~\citep{yangTaxonomizingLocalGlobal2021}. 
Interestingly, the phases and phase transitions can be linked to the structure of the loss landscape~\citep{yangTaxonomizingLocalGlobal2021}, especially its global structure. We note that the study of global loss landscapes has been an active area of research, and it is generally understood that the properties of neural networks cannot be fully captured by local sharpness alone~\citep{fortLargeScaleStructure2019, fort2020deep}. \citet{yangTaxonomizingLocalGlobal2021} present the first empirical attempt to quantify the transition from a globally well-connected to a globally less well-connected loss landscape. Specifically, metrics such as the \emph{training} loss, the sharpness of local minima, and mode connectivity or representation similarity computed on the training data can be used to identify the phase of a model.

\paragraph{Loss Landscape Metrics.}
~\citet{yangTaxonomizingLocalGlobal2021} categorize phases in load-temperature variations using four metrics. The first metric is the training loss, which evaluates whether the training data is interpolated. The other metrics describe the sharpness of the local minima, the similarity between models trained using different random seeds, and the connectivity between different local minima of the loss landscape. 
It should be noted that~\citet{yangTaxonomizingLocalGlobal2021} used a certain set of metrics to measure these loss landscape properties, but there are alternative metrics available. For example, the sharpness of local minima can be measured using \emph{adaptive sharpness metrics}~\citep{andriushchenko2023modern,kwon2021asam},
while similarity can be measured using \emph{disagreement}~\citep{theisenWhenAreEnsembles2023}.\looseness-1

We define the loss landscape metrics following~\citet{yangTaxonomizingLocalGlobal2021}. Let $\vtheta \in \mathbb{R}^m$ denote the learnable weight parameter, and let $\mathcal{L}$ be the loss function. We compute metrics using the train set unless stated otherwise.\looseness-1

\textbf{Hessian-based Metrics.}
The Hessian matrix $\mathbf{H}$ at a given point $\vtheta$ can be represented as $\nabla^2_\mathbf{\vtheta} \mathcal{L}(\vtheta) \in \mathbb{R}^{m \times m}$. The largest eigenvalue $\lambda_{\max}(\rmH)$ and trace $\text{Tr}(\mathbf{H})$ are used to summarize the local curvature properties in a single value. Specifically, a larger value of the top eigenvalue or trace indicates greater sharpness.\looseness-1

\newpage

\textbf{Mode Connectivity.}
The mode connectivity assesses the presence of low-loss paths between different local minima and reflects how well different solutions are connected in the parameter space, indicating smoother and more generalizable loss landscapes. It is common to fit Bézier curves $(\gamma_{\phi}(t)$ between two models $\vtheta$ and $\vtheta'$, and subsequently compute mode connectivity $\texttt{mc}$ as
\[
\texttt{mc}(\vtheta, \vtheta') = \frac{1}{2}(\mathcal{L}(\vtheta) + \mathcal{L}(\vtheta')) - \mathcal{L}(\gamma_{\phi}(t^*)),
\]
where $t^* = \underset{t}{\mathrm{argmin}} \left| \frac{1}{2}(\mathcal{L}(\vtheta) + \mathcal{L}(\vtheta')) - \mathcal{L}(\gamma_{\phi}(t)) \right|$. 
Here, $\texttt{mc}<0$ indicates a loss barrier between the two models and hence poor connectivity.
$\texttt{mc}>0$ reveals lower loss regions between the models which indicates poor training.
$\texttt{mc}\approx0$ indicates well-connected models.

\textbf{CKA Similarity.} Centered Kernel Alignment (CKA)~\citep{kornblithSimilarityNeuralNetwork2019} is used to evaluate the similarity between representations learned by different NNs, providing a measure of consistency and robustness in feature learning.
CKA helps to understand how similar the learned representations are across different minima, linking representation similarity to landscape structure and generalization performance. 
The CKA between output logits $\rmX$ and $\rmY$ generated by $\vtheta$ and $\vtheta'$ is computed as
 \[
\texttt{cka} = \frac{\text{HSIC}(\rmK, \rmL)}{\sqrt{\text{HSIC}(\rmK, \rmK) \cdot \text{HSIC}(\rmL, \rmL)}}
 \]
 where $\text{HSIC}$ is the Hilbert-Schmidt Independence Criterion and $\rmK$ and $\rmL$ are the Gram matrices of $\rmX$ and $\rmY$, respectively.

\begin{wrapfigure}{r}{0.45\textwidth}
\centering
\vspace{0mm}
\includegraphics[width=1.0\linewidth,keepaspectratio]{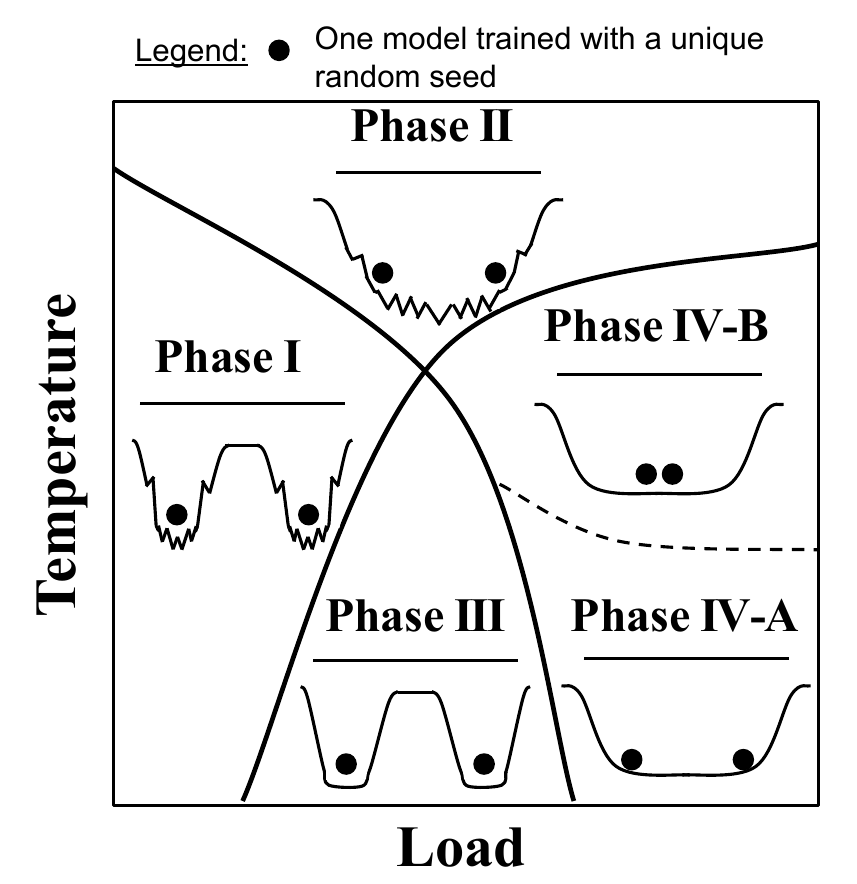}
\caption{Five-phase taxonomy in NN hyperparameter space~\citep{yangTaxonomizingLocalGlobal2021}, varied by load-like and temperature-like parameters. Our zoos cover all five phases.\looseness-1
}\label{fig:phase-caricature}
\vspace{5mm}
\end{wrapfigure}
\paragraph{Phase Taxonomy.} Based on loss landscape metrics, the NN hyperparameter space is divided into five distinct phases, as depicted in Figure~\ref{fig:phase-caricature}.
\begin{itemize}[leftmargin=*, nosep, noitemsep] %
    \item \textbf{Phase I (underfitted \& undersized)}: Train loss is high; Hessian metrics are relatively large (indicated by a rugged basin); Mode connectivity is negative (indicated by a barrier between two local minima). 
    \item \textbf{Phase II (underfitted)}: Train loss is high; Hessian metrics are relatively large; Mode connectivity is positive. 
    \item \textbf{Phase III (undersized)}: Train loss is small; Hessian metrics are relatively small (smooth basin); Mode connectivity is negative.\looseness-1
    \item \textbf{Phase IV-A (overfitted)}: Train loss is small; Hessian metrics are relatively small; Mode connectivity is near-zero; CKA similarity is relatively small.\looseness-1
    \item \textbf{Phase IV-B (generalizing)}: Train loss is small; Hessian metrics are relatively small; Mode connectivity is near-zero (no barrier between minima); CKA similarity is relatively large.
    \looseness-1
\end{itemize}

\citet{yangTaxonomizingLocalGlobal2021} use these metrics, along with fixed thresholds, to partition training into distinct phases. Training loss separates early phases (Phase I and II) from later ones (Phase III and IV). Mode connectivity, depending on whether it is positive or negative, differentiates Phase I from Phase II. Hessian trace and CKA similarity are then used to further divide Phase IV into two sub-phases. \citet{zhou2024MD} formalize this procedure as a hierarchical decision tree, where thresholds for each metric can be learned using a small number of reference models. This optimization is performed using the bounded Brent method to maximize prediction accuracy on ‘failure modes’ of models. They show that the resulting phase boundaries can be transferred effectively to new tasks or model architectures.


\section{Phase Transition Model Zoos}
\label{sec:model_zoo_gen}
To create a population of models that covers relevant phases and can be used to evaluate for phase transitions, 
we train structured populations of NNs with several architectures on different datasets following the blueprint introduced by~\citet{unterthinerPredictingNeuralNetwork2020} and~\citet{schurholtModelZoosDataset2022}. Within each \textit{model zoo} population, we systematically vary load-like and temperature-like hyperparameters to realize all of the phases.
For every model in the zoo, our dataset includes multiple checkpoints (i.e. saved model weights), at different training epochs.
We annotate these samples with performance metrics (training and test loss and accuracy), as well as the loss landscape metrics outlined in Section~\ref{sec:loss-landscape}. 
Since the exact localization of phase boundaries is still a topic for ongoing research, we annotate each model with its loss landscape metrics as quantitative ground truth, from which different methods can derive slightly different phase transition boundaries.
We further track loss and accuracy on train and --- if available --- validation and test sets.
In the following, we first detail our model zoo generation scheme, which we use to create 12 zoos on three domains: computer vision, natural language and physical systems. Further details can be found in Appendix \ref{app:model_zoo_gen}. Subsequently, we analyze our models with conventional performance metrics, but also with loss landscape metrics to quantify the qualitative diversity of the proposed zoos and validate that all of the phases are realized.

\paragraph{Computer Vision Model Zoos.}
Ten computer vision zoos form the foundation of our dataset and demonstrate the generalization of the phase transitions accross architectures and datasets, while the language and SciML zoos show generalization of the phase transitions to other domains and tasks. 
We generate the computer vision zoos from combinations between two architectures $\{\textrm{ResNet}, \textrm{ViT}\}$ of different sizes and four standard computer vision datasets $\{\textrm{SVHN}, \textrm{CIFAR-10}, \textrm{CIFAR-100},$ $\textrm{TinyImagenet}\}$~\citep{svhn, cifar, tinyimagenet}. Details on the model zoo configurations can be found in Table \ref{tab:hyperparams_vision} in the Appendix.
We choose ResNet~\citep{heDeepResidualLearning2016} and ViT~\citep{dosovitskiyImageWorth16x162021} architectures for the zoos because of their proliferation in computer vision to achieve representative populations. Importantly, ResNet and ViT architectures allow smooth scaling of model width and thus model capacity for the same architecture without the need to adjust the learning scheme. 
Similarly, the set of datasets evaluate generalization accross different data distributions.
Taken together, the vision zoos contain $\sim$ 1.8k unique models and $\sim$ 55k checkpoints. \looseness-1

\newpage

\paragraph{Large Language Model Zoo.}
For the natural language zoo, we train GPT-2~\citep{radfordLanguageModelsAre2019} models, following the reference implementation in~\citep{karpathyKarpathyNanoGPT2025}. We use decoder-transformer GPT-2 to include coverage of modern LLMs in our dataset, which have since largely converged to decoder-only transformers. The GPT-2 architecture can be scaled just like the ViTs by changing the width, allowing smooth variations of the model capacity. We train on openwebtext which replicates the original dataset of GPT-2~\citep{Gokaslan2019OpenWeb}.  
We use Chinchilla scaling laws to determine the reference ratio of a GPT-2 nano model to data and train on 2.29B tokens~\citep{hoffmannTrainingComputeOptimalLarge2022}. Notably, work on scaling laws established relations between model size and data amount for compute-optimal performance. This suggests that phase distributions identified on GPT-2 nano-sized models generalize to much larger models if the data amount is scaled accordingly, within the range of the scaling laws. This significantly increases the expressiveness of the zoo. Interestingly, this is exactly what the relative load parameter as the ratio of model capacity to data suggests, and what makes the phase layout on these abstract axes powerful. In its final grid, the language zoo contains 264 unique GPT-2 models with $\sim$ 1.3k checkpoints.\looseness-1

\begin{figure*}[t]
\vspace{-12pt}
\centering
\includegraphics[width=0.99\linewidth]{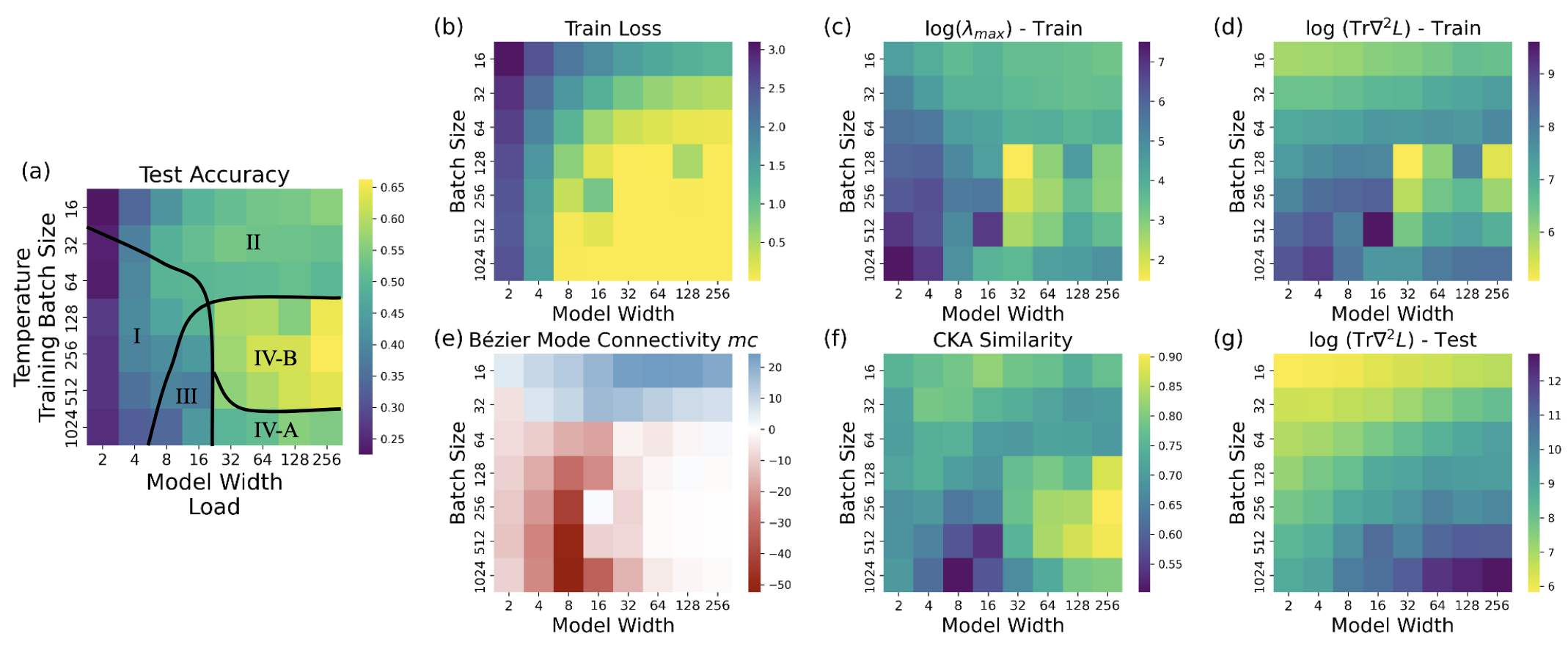}
\caption{
Performance and loss landscape metrics for the CIFAR-100 ResNet-18 model zoo. \textbf{(a):} test accuracy and phases of models in the zoo. \textbf{(b):} training loss; \textbf{(c-g)} different loss landscape metrics introduced in Section~\ref{sec:loss-landscape}. Our model zoos cover all phases identified in previous work~\citep{yangTaxonomizingLocalGlobal2021}.
\looseness-1
}
\label{fig:loss_landscape_cifar100_resnet18}
\vspace{-8pt}
\end{figure*}

\paragraph{SciML Model Zoo.}
Lastly, we also create zoos of models for the simulation of physical systems. With that, we want to evaluate whether phases and phase transitions generalize even beyond the conventional deep-learning domains.
Specifically, we train physics-informed neural networks (PINNs)~\citep{raissiPhysicsinformedNeuralNetworks2019} to learn the solution to partial differential equations, following the implementation setup in \citep{krishnapriyan2021characterizing}.
To curate the dataset, we simulate the 1D convection partial differential equation (pde) and randomly sample the domain's collocation points (position, time) as the data samples.
Each data sample is comprised of position and time as input and the PDE solution value as the label.
The training objective includes minimizing the loss of data prediction, boundary/initial conditions, and physics-based regularization.
We train a 4-layer fully connected NN with 50 neurons per layer and tangent activation function using the L-BFGS optimizer.
By varying temperature and load paramters, we train a zoo with $700$ unique PINNs and $7\,000$ checkpoints.\looseness-1

\paragraph{Load and Temperature Variations.}
For all model zoos on all domains, we introduce specific variations in the training hyperparameters to obtain models in all phases. Previous work identifies the phases on the surface spanned by load-like and temperature-like hyperparameters~\citep{martinRethinkingGeneralizationRequires2019,yangTaxonomizingLocalGlobal2021}
The load-like parameters can be understood as the quantity and/or quality of data relative to the model capacity. 
Temperature represents the noisiness of the training process. 
Following previous work, we realize variations in load by changing the model width. Increasing the model width increases model capacity and thus decreases the relative load. By varying the width, we achieve variations in model capacity without changing the architecture or having to adapt the training scheme. 
In ResNets, the width directly refers to the number of channels. 
In transformer models (ViTs and GPT-2) , we realize width by changing the \texttt{model\_dim} parameter, i.e. the size of intermediate representations. 
In SciML models, we change the load by varying the PDE convection coefficient, as it changes the complexity of the data.
To realize variations in temperature, we choose to adapt the batch size or learning rate. 
Here, lower batch size and higher learning rate increase the noisiness of the training updates and this increases the temperature. 
For every combination on the grid, we train three different models using three random seeds.
All other hyperparameters are kept constant between the models.

\section{Empirical Evaluation of Phases in the Model Zoos}
\label{sec:phases_empirics}
The model zoos are designed to cover different phases. In the following, we validate phase coverage by testing for the phases introduced by~\citet{yangTaxonomizingLocalGlobal2021} summarized in Section~\ref{sec:loss-landscape}. 
Full phase plots for all 12 zoos and further details can be found in Appendix \ref{app:zoo_eval}.

\subsection{Phases Systematically Emerge Across all Zoos}
Our experiments demonstrate that phase transitions are consistently present in the training of neural networks across all domains, architectures, and datasets evaluated. The exact phase layout is affected by the architecture, dataset, and data augmentation strategies. The specific characteristics of these phases remain consistent with the four-phase taxonomy outlined in previous studies, validating our experiment setup.

A central reason we focus on load-like and temperature-like parameters is that they unify the effects of many hyperparameters into two axes. Empirically, models occupying the same phase on these axes exhibit qualitatively similar properties—regardless of which specific training factors (e.g., batch size, learning rate, or data size) gave rise to those load and temperature values. Consequently, rather than exhaustively exploring every hyperparameter combination, we can vary these two axes to capture all known phases. This is not only supported by prior work—where phase boundaries remain stable across different load–temperature choices (Appendix D in \citet{yangTaxonomizingLocalGlobal2021})—but also reflected in our results, suggesting strong generality of phase transitions to a broad range of architectures and training regimens.

\begin{figure}[t]
    \vspace{-12pt}
    \centering
    \includegraphics[width=0.90\textwidth]{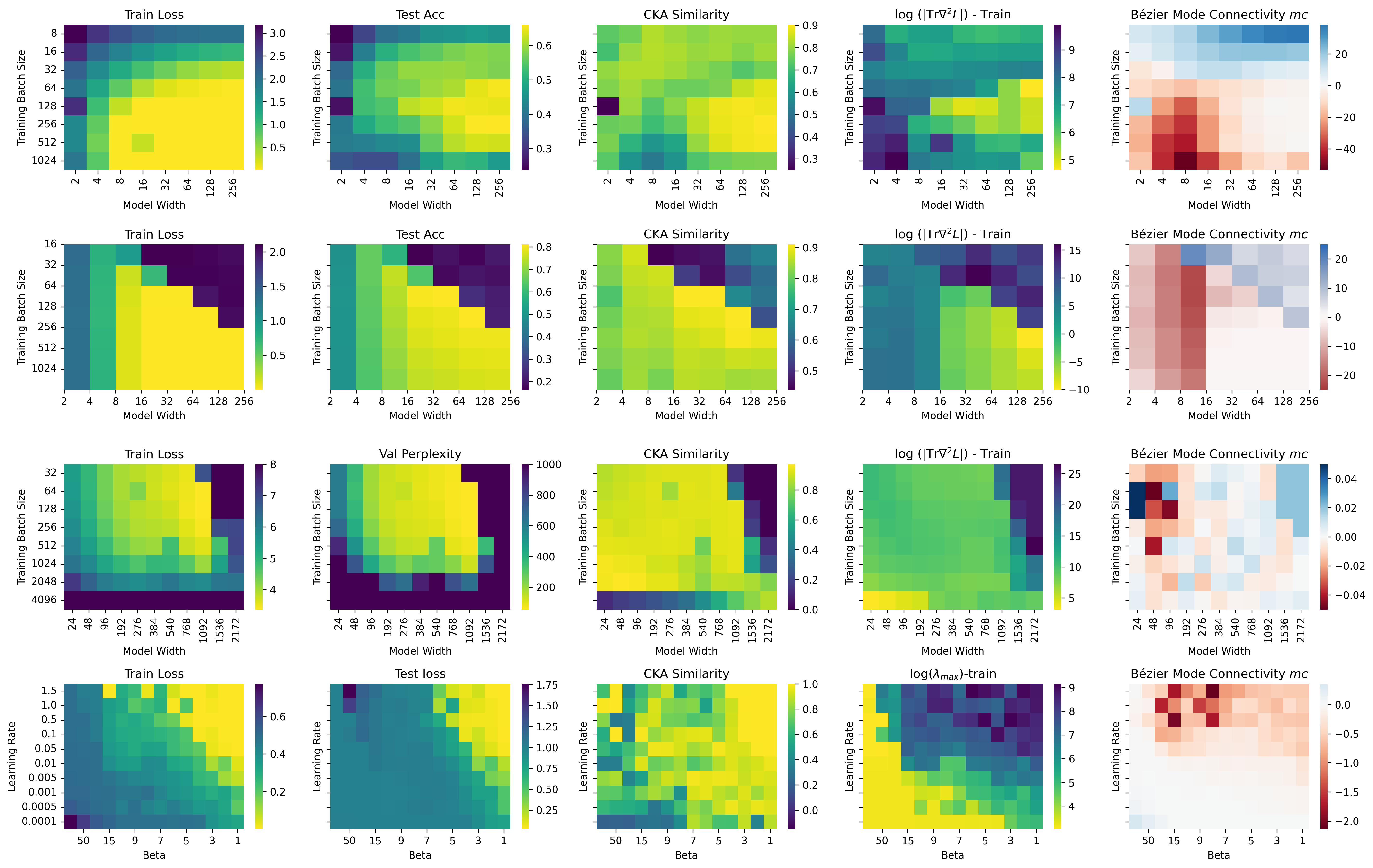}
    \caption{Phase plots accross different architectures, domains and tasks. Phases and phase transitions emerge in train loss, test performance, and loss landscape metrics, across different domains, tasks, and architectures. The exact layout is determined by the architecture, domain, and task. For example, transformers (ViT and GPT-2) have a sharp transition in the high-temperature low-load (top right) regime, which ResNets and PINNs do not show.}
    \label{fig:domain_generalization}
\end{figure}

\paragraph{Phases Generalize Computer Vision Architectures, Model Sizes and Datasets.}
As illustrated in Figure~\ref{fig:loss_landscape_cifar100_resnet18} and in Figures \ref{fig:phase_plot_r18_c10}-\ref{fig:phase_plot_sciml}, the phases manifest clearly in the combination of loss landscape metrics such as Hessian trace, mode connectivity, and CKA similarity. 
While different methods to locate phase transitions may identify slightly differ boundaries, the evaluation of the performance and loss landscape metrics in the phase plots clearly demonstrates the coverage of the phases described in previous work.
In particular, Phase IV-B, associated with the best test accuracy, is marked by low loss and high generalization performance. On the ResNet zoos, our results reveal that learning rate decay plays a significant role in shaping the phase distribution. Specifically, decaying the learning rate by $1e4$ under cosine annealing increases the area of Phase IV (well-trained regime) while reducing the presence of Phase II (under-trained regime), as the effect of batch size variations diminishes. This may be an indication of why learning rate decay is so successful.
Our experiments show that the phase transitions generalize across different datasets, architectures, and training regimes.
ViTs, for instance, display a sharper transition from Phase II to IV when trained without heavy augmentation (Figure \ref{fig:phase_plot_vit_c10}). GPT-2 models exhibit a comparable pattern (Figure \ref{fig:phase_plot_gpt2_openwebtext}), suggesting that phase layouts remain consistent across distinct architectures, including transformers.

\paragraph{Phases Generalize Across Domains.} 
Notably, the phases and phase transitions on the load-temperature landscape emerge on vastly different domains, from computer vision and classification to natural language and generation, and even physics and simulation, see Figure \ref{fig:domain_generalization} and Figures \ref{fig:phase_plot_r18_c10}-\ref{fig:phase_plot_sciml} in the Appendix.
While the exact layout varies between the zoos, and is affected by architecture and task, our results suggest that the fundamental drivers of phases translate directly. 
Accross all domains and architectures, the high test performance region is contained in the region with low train loss. However, the low train loss region is significantly larger.
As described in previous work, loss landscape metrics can be used to identify the layout of best test performance. The combination of high CKA similarity, low curvature (hessian trace) and low mode connectivity error computed on the training data predict high test performance. 
There are two exceptions that bear further investigation: 
First, in SciML models, while the phase layout aligns with previous work~\citep{geniesse2024visualizing}, the high performance phase has high hessian traces. The rason for that may lie in the training problem of PINNs, which is known to have particular properties~\citep{cheng2024physics,krishnapriyan2021characterizing}. In our zoos, low learning rates may cause PINNs to get trapped in flat local minima, while global minima are sharp.
Second, mode connectivity on the GPT-2 models is computed on the loss rather than on the prediction error which adds noise, due to the high dimensionality of the classification problem. \looseness-1

These experiments suggest that since the phases generalize to different domains and architectures, findings of specific phases in specific applications on one domain can also be found in all domains. 
While some phase transitions on non-CV domains have been described before, the direct match is to the best of our knowledge novel and underscores the expressiveness of phases and its usefulness for the composition of diverse zoos.
By confirming the presence of all known phases, we establish that this dataset is not only a tool for studying phase transitions but also a resource for designing and testing phase-aware training algorithms. Next, we turn to downstream methods to illustrate why phases matter post-training.\looseness-1

\subsection{Phase Transitions in Neural Network Methods}

\begin{figure}[t]
    \vspace{-12pt}
    \centering
    \includegraphics[width=0.95\textwidth]{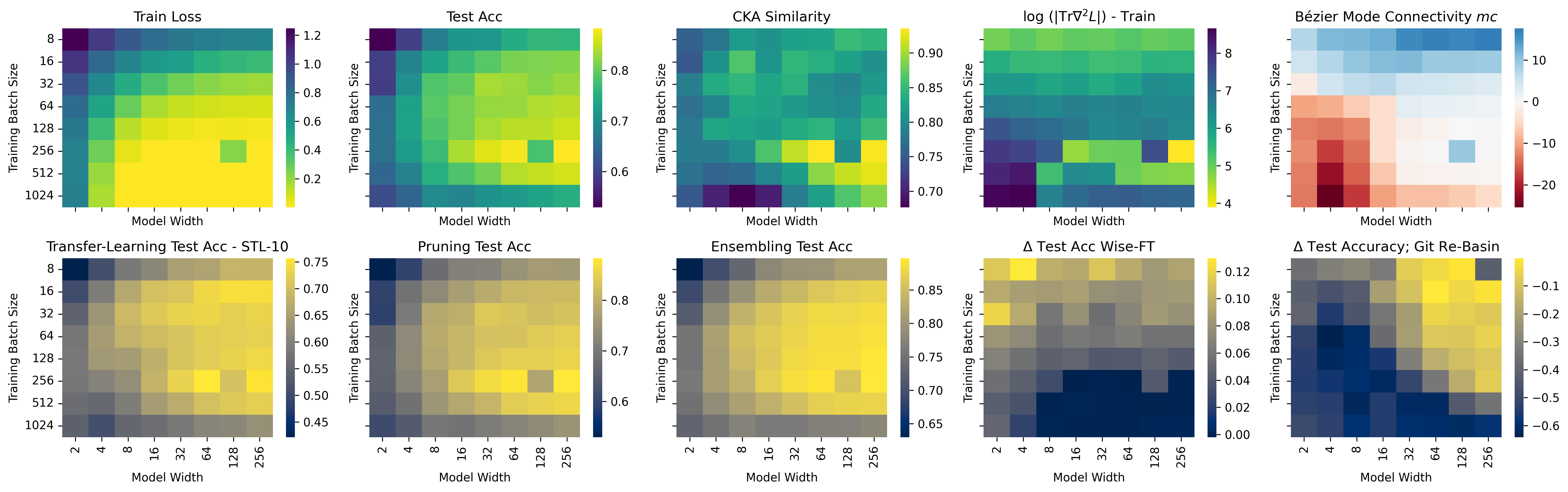}
    \caption{Phases and phase transitions of ResNet-18 models trained on CIFAR-10 in \textbf{top:} pre-training and loss-landscape metrics and \textbf{bottom:} downstream methods. Phase transitions emerge in downstream methods like fine-tuning, pruning and weight averaging. We use a different color palette for the downstream methods figures in order to make them easily identifiable.}
    \label{fig:phases_downstreamtasks}
    \vspace{-12pt}
\end{figure}

Building on the observation that phase coverage describes performance and generalization during training, we now examine how phases influence four popular downstream methods: fine-tuning, pruning, ensembling, and weight averaging.
%
%
In practice, many widely‐used methods rely on \emph{pre‐trained} models—raising the question of whether and how a model’s phase continues to matter when its weights are reused. Below, we evaluate four canonical techniques and discuss how our phase‐covering model zoos enable a more comprehensive analysis of their properties. Across all these downstream applications, we find distinct phases in downstream performance, see Figure \ref{fig:phases_downstreamtasks}. Some of them overlap with the (pre-)training phases, some are distinct. These results are notable for three reasons: 
\textbf{(i)} it demonstrates that phase transitions exist broadly 
\textbf{(ii)} the load-temperature axis of our zoos is general enough to cover phase transitions in several downstream applications, beyond pre-training, significantly improving the applicability of our zoos;
\textbf{(iii)} the datasets facilitate research of \emph{where} these methods work or fail beyond point-wise evaluation, and build towards phase‐aware method design.

\paragraph{Fine-Tuning and Transfer Learning.}
When fine‐tuning, e.g.\ a CIFAR10‐pretrained ResNet-18 on STL‐10 (Figures \ref{fig:phase_plot_r18_c10},\ref{fig:phase_plot_r18_ti}, bottom left), we find that networks ending in a well‐trained (Phase IV) regime yield stronger adaptation. Conversely, under‐fitted or over‐fitted phases struggle to reach comparable accuracy despite identical fine‐tuning settings. This aligns with earlier observations that good phases not only benefit the original task but also confer more adaptable representations. For less aligned tasks like Tiny-Imagenet to STL-10 (Figure \ref{fig:phase_plot_r50_ti}), we notice a phase shift from pre-trained to fine-tuned models. Understanding the relation between pre-training tasks and phase alignment using our model zoos can help pre-train or choose models to fine-tune in a more targeted fashion. 

\paragraph{Pruning.}
Pruning has been shown to exhibit phase‐dependent outcomes~\citep{zhou2023three}. We confirm that Phase IV networks often preserve accuracy better under high sparsity, while certain under‐fitted/low‐connectivity phases degrade quickly (Figure \ref{fig:phases_downstreamtasks} bottom, second from left). By comparing across load‐temperature grids, our model zoos help pinpoint where and why pruning works best. 

\paragraph{Ensembling.}
Combining model predictions (ensembling) has been investigated for phase transitions~\citep{theisenWhenAreEnsembles2023}. Our experiments confirm that ensembles tend to improve performance in load‐temperature regions where local minima are well‐connected (Figure \ref{fig:phases_downstreamtasks} and \ref{fig:phase_plot_vit_c10}). Notably, the phase of high-performance ensembles is larger than the individual model high-performance Phase IV, demonstrating the robustness benefit of ensembling with respect to pretraining settings. In practice, that corresponds to phases with near‐zero mode connectivity barriers. Where mode connectivity is negative, ensembling provides less benefit.
The zoo’s phase annotations thus indicate not just whether ensembling helps, but \emph{which} configurations best exploit ensemble diversity.

\paragraph{Weight Averaging.}
Averaging weights has recently gained popularity to improve robustness or combine knowledge~\citep{izmailovAveragingWeightsLeads2019a,wortsmanModelSoupsAveraging2022,ilharcoEditingModelsTask2022}.
We test averaging across training epochs (Wise-FT~\citep{wortsmanRobustFinetuningZeroshot2022}) and across pre-training seed (git re-basin~\citep{ainsworthGitReBasinMerging2022}).
Our experiments (Figures~\ref{fig:phase_plot_r18_c10}--\ref{fig:phase_plot_r50_svhn}) reveal distinct phase‐dependent outcomes: Wise‐FT is notably effective in high‐temperature regions (Phases I or II) where averaging reduces noisy training updates, while in flatter phases (e.g., IV) it can yield little additional benefit. Git Re‐Basin, on the other hand, hinges on positive mode connectivity; negative‐connectivity phases (I or III) remain difficult to merge without performance loss. Hence, Hessian trace and mode connectivity strongly govern weight‐averaging success, underscoring the value of our phase‐annotated model zoos for pinpointing when and where averaging is most fruitful. Here in particular, point-evaluations of merging a few models lack the big picture necessary to build understanding.

\section{Model Zoo Applications in Weight Space Learning}
\label{sec:app_methods}

The ``weight space'' perspective treats trained NN weights as a data modality in its own right, enabling various novel applications such as model analysis, merging, editing, generation etc. Most of these approaches require a training set of model weights, as well as structured, diverse evaluation sets.
Our phase-annotated model zoos directly cater to this need: they provide large, systematically varied populations of trained models with a quantifiable notion of diversity, alongside key loss-landscape metrics. They can be used to train weight-space models and therefore can facilitate the development of future methods in the area. They can also be used as a diverse, structured, standardized evaluation set for such methods. In the following, we highlight recent developments in weight space learning and illustrate how our dataset can contribute to research in those areas.

\subsection{Learning Dynamics Analysis}

In this paper, we focus on the phase transition phenomenon as studied in the statistical physics literature~\citep{engel2001statistical}. This line of work links physics models, such as the Curie-Weiss model of magnetization, to problems in statistical inference. It often uses strong analytical tools like the replica method to compute free energy and characterize the phase transitions of learning systems. For a comprehensive overview, see \citet{zdeborova2016statistical}. At the same time, there is a growing body of research on phase transitions in learning dynamics. These studies show that the behavior of learning dynamics/trajectories can shift sharply as time evolves or as hyperparameters change. Some of this work also connects to statistical physics, while others develop independently. For example, \citet{lewkowycz2020large} identify three learning phases under different initial learning rates: lazy, catapult, and divergent. They show that a large but non-divergent learning rate yields the best test performance. As another example, \citet{baity2018comparing} describe three stages of training: exploration of the loss landscape, aging dynamics with an increasing number of flat directions, and a final phase resembling diffusion. Extending our analysis to capture more of these phenomena in various learning dynamics is an important direction for future work.

\subsection{Model Training}
A core challenge in deep learning is hyperparameter tuning, which often relies solely on validation performance~\citep{jaderberg2017population,li2020system}. 
By shifting toward a weight space perspective, practitioners can leverage loss-landscape signals—e.g., Hessian trace, mode connectivity—to guide models into more favorable phases (e.g., flatter minima). 
Recent works demonstrate that such information can enable faster or more reliable optimization~\citep{yao2018hessian, zhou2023three, zhou2024MD}. Lack of availability of specialized model zoos with loss landscape metrics however limit their evaluation to few model zoos.
Our model zoos address this limitation, as they contain checkpoints covering model trainings in length, and are enriched with corresponding loss landscape metrics. With our broad coverage of the load-temperature axes, as well as multiple data modalities, our dataset offers opportunities to study the influence of these parameters on the training of the models in a variety of contexts and test methods across different domains and architectures.

We demonstrate the potential of our dataset for phase-aware training algorithms with a simple hyper-parameter optimization experiment to maximize performance gains in a single step. We consider any state on the grid, and attempt to change temperature or load like parameters to improve performance. That is, for each configuration, we perform one-step hyperparameter optimization (to any other load-temperature cell) and measure the resulting performance improvement. As stand-in for validation-based methods, we use a random search that keeps the post-hoc better cell in validation metrics. We compare against a simple decision rule based on the current loss landscape metrics.

\textbf{Random Search Baseline}: For each model configuration, we randomly select one of three tuning actions: increase model width, increase batch size, or decrease batch size. The magnitude of adjustment is also randomly sampled from a predefined range 1-5.

\textbf{Loss Landscape-Guided Search}: For each model configuration, we first classify its phase using four loss landscape metrics and the phase boundaries by identifying thresholds similarly to \citet{yangTaxonomizingLocalGlobal2021}. We then apply deterministic tuning directions:
\begin{itemize}
    \item Phase I and III: Increase model width
    \item Phase II: Increase batch size
    \item Phase IV-A: Decrease batch size
    \item Phase IV-B: No parameter adjustment
\end{itemize}

While tuning directions are deterministic based on phase classification, tuning magnitudes are randomly sampled to maintain a fair comparison with the baseline. Please note that this one-step procedure can easily be extended to multi step optimization. We focus on single step solely for evaluation purposes. To ensure statistical reliability, we repeat each experiment 50 times and report both mean performance gains and standard deviations. We report results in Table~\ref{tab:phase_aware_training}.

\begin{table}
\captionof{table}{
Evaluating the performance of phase-aware hyperparameters optimization approaches compared to a random search baseline, on ResNet-18 models. We report average over 50 runs as well as standard deviation. For all three model zoos evaluated, phase-aware hyperparameters optimization outperforms the baseline.
}
\setlength{\tabcolsep}{5pt}
\label{tab:phase_aware_training}
\centering
\small
\begin{tabularx}{0.9\textwidth}{lccc}
\toprule
\textbf{Algorithm} & \textbf{CIFAR-10} & \textbf{CIFAR-100} & \textbf{TinyImageNet} \\
\cmidrule(lr){1-1}\cmidrule(lr){2-2}\cmidrule(lr){3-3}\cmidrule(lr){4-4}
\textbf{Random Search} & $0.012 \pm 0.006$ & $0.026 \pm 0.009$ & $0.037 \pm 0.012$ \\
\textbf{Phase-Aware Optimization} & $\mathbf{0.038 \pm 0.004}$ & $\mathbf{0.072 \pm 0.002}$ & $\mathbf{0.116 \pm 0.007}$ \\
\bottomrule
\end{tabularx}
\end{table}

The results clearly show that using loss landscape information to optimize model hyperparameters significantly improves performance and efficiency of a single optimization step. We would further like to refer the reader to \citet{zhou2024MD}, who have systematically evaluated the directional benefit of loss landscape metrics for hyper-parameter optimization over conventional validation metrics. 

\newpage

\subsection{Model Analysis}
Weight spaces can also be leveraged to infer model properties --- test accuracy~\citep{eilertsenClassifyingClassifierDissecting2020, unterthinerPredictingNeuralNetwork2020}, generalization power~\citep{schurholtHyperRepresentationsGenerativeModels2022}, backdoor presence~\citep{langosco2023detecting}, task~\citep{herrmannLearningUsefulRepresentations2024}, INR class~\citep{deluigiDeepLearningImplicit2023, navonEquivariantArchitecturesLearning2023}, relations to other models~\citep{horwitzOriginLlamasModel2024} etc. --- directly from the model weights. Most of these works train their own model zoos, limiting the possibility to effectively compare their performance, and making the evaluation on a wide variety of training parameters and modalities prohitively expensive.

Our dataset offers a broad distribution of architectures, training states, and documented loss-landscape metrics, creating an ideal environment for developing or benchmarking weight space predictors, in particular to predict tasks, hyperparameters, performance and loss landscape metrics. Because our zoos deliberately include ``bad'' phases (I, III) as well as ``good'' phases (IV), methods that learn from this data will be more robust across real-world scenarios. We demonstrate the usability of our dataset by conducting a small pilot experiment following~\citet{unterthinerPredictingNeuralNetwork2020}: we extract simple weight statistics (e.g.\ per-layer means and quintiles) for each model, then use a linear probe to predict both performance and loss-landscape metrics. Table~\ref{tab:metrics_prediction} reports the $R^2$ values on a held-out test split, suggesting that curvature (log Hessian) and representation similarity (CKA) are partly learnable from raw weights. Even mode connectivity (MC)---which depends on pairs of models---shows signs of predictability, albeit less strongly. 

\begin{wraptable}{r}{0.6\linewidth}
\vspace{-12pt}
\captionof{table}{
Predicting ResNet-18 performance and loss-landscape properties from raw weight statistics: $R^2$ on unseen test sets. 
}
\setlength{\tabcolsep}{3pt}
\label{tab:metrics_prediction}
\centering
\small
\begin{tabularx}{1.0\linewidth}{cccccc}
\toprule
\textbf{Data} & \textbf{Test Acc} & \textbf{GGap} & \textbf{CKA} & \textbf{log(Tr$\nabla^2 L$)} & \textbf{MC} \\
\cmidrule(lr){1-1}\cmidrule(lr){2-2}\cmidrule(lr){3-3}\cmidrule(lr){4-4}
\cmidrule(lr){5-5}\cmidrule(lr){6-6}
SVHN  & 0.92 & 0.67 & 0.41 &  0.48 & 0.68 \\
CIFAR10 & 0.54 & 0.53 & 0.46  & 0.61 & 0.81 \\
CIFAR100 & 0.89 & 0.67 & 0.13  & 0.79 & -2.93 \\
TinyImgNet & 0.90 & 0.63 & 0.28  & 0.72 & 0.52 \\
\bottomrule
\end{tabularx}
\vspace{-2mm}
\end{wraptable}

These preliminary findings underscore two key insights. First, the metrics that define each \emph{phase} appear sufficiently structured that simple regression models can partially recover them from the weights alone. Second, this partial predictability could serve as an efficient \emph{guidance signal} during training or hyperparameter optimization, instead of computing expensive Hessian or connectivity measures. While more advanced predictors will be needed to achieve high fidelity, our results highlight how phase-annotated model zoos open the door for practical methods that leverage \emph{weight-space learning} to navigate the loss landscape. 

\subsection{Model Editing} Working on the weights of NNs as a modality opens up opportunities to directly alter them. Existing applications include re-aligning models with regard to permutation symmetries~\citep{ainsworthGitReBasinMerging2022, navonEquivariantDeepWeight2024}, domain adaptation~\citep{navonEquivariantArchitecturesLearning2023}, knowledge unlearning~\citep{meng2023mass}, or model merging~\citep{ilharcoEditingModelsTask2022}. Here again, the wide variety of models in our zoos along hyperparameters, architectures and modalities will allow easy development and standardized evaluation of such methods using phase information signals.

\subsection{Model Generation}
Finally, there is a growing push to \emph{generate} NN weights, either to initialize new models without full retraining or to combine multiple networks for improved performance~\citep{schurholtHyperRepresentationsGenerativeModels2022,peeblesLearningLearnGenerative2022, schurholtScalableVersatileWeight2024,wang2024recurrent,putterman2024learninglorasglequivariantprocessing,soro2025diffusionbased, meynent2025structure,falk2025impact}. However, as we argue above, existing datasets to that end either contain only well-trained models or do not control for diversity in a quantifiable way. Our dataset provides precisely the sort of phase-diverse, architecture-diverse examples that generative models can learn from. 
By conditioning on quantifiable quality metrics that describe the phase, or on loss landscape information like curvature or connectivity, a generator trained on our zoos could be trained to produce custom weights with predictable properties.\looseness-1

\section{Discussion}

\paragraph{Limitations}

While our model zoos cover the domains of vision, language and science, most Phase Transition Model Zoos are composed of classification models in the computer vision domain. 
We chose to explore phase transitions in vision exhaustively, and to demonstrate the generalization of phases as a concept with the additional zoos in language and science.
Our work presents itself as a first step to making model zoos comprehensively cover phase transitions for a variety of applications, and we leave its extension to further tasks and domains for future work.\looseness-1

\paragraph{Conclusion}
The Phase Transition Model Zoos represent the largest structured collection of models across architectures, tasks and domains annotated with detailed loss landscape metrics. With it, we provide the research community with a powerful dataset for weight space learning, and a tool to evaluate neural network performance across different phases. 
By systematically covering phase transitions, it allows the study of robustness, generalization, and failure modes of deep learning methods in a much more nuanced, comprehensive, and reliable way.\looseness-1
We demonstrate the relevance of phase transitions by identifying phases in methods like fine-tuning, transfer learning, pruning, ensembling, and weight averaging. We show that these phases significantly affect performance and that their impact varies from one method to another, indicating valuable insights that can be gained with our dataset beyond conventional performance metrics.\looseness-1
%

\impact{Our work introduces a dataset of neural networks that systematically spans multiple phases of training across different architectures and domains. By providing controlled diversity in the weight space, we aim to accelerate research on weight space learning, a growing field that treats trained weights as a data modality. This dataset can help researchers analyze when and how different training phases arise, develop phase-aware training and downstream methods, and gain deeper insights into the fundamental properties of neural networks. \looseness-1

This project inherits the typical risks of large-scale machine learning research and progress of the field. While we hope our dataset can save ressources in phase aware research, the computational resources for creating, analyzing, and extending such model zoos can be significant.\looseness-1
}

\acks{KS, LM and DB are supported by the Swiss National Science Foundation research project grant 10001118 and SPRIND through the ModelFoundry project. 
YZ and YY are supported by DOE under Award Number DE-SC0025584 and funding from Dartmouth College.}

\vskip 0.2in
\bibliography{./bib_auto.bib, ./bib_manual.bib, ./bibliography.bib}

\begin{thebibliography}{105}
\providecommand{\natexlab}[1]{#1}
\providecommand{\url}[1]{\texttt{#1}}
\expandafter\ifx\csname urlstyle\endcsname\relax
  \providecommand{\doi}[1]{doi: #1}\else
  \providecommand{\doi}{doi: \begingroup \urlstyle{rm}\Url}\fi

\bibitem[Ainsworth et~al.(2022)Ainsworth, Hayase, and Srinivasa]{ainsworthGitReBasinMerging2022}
Samuel~K. Ainsworth, Jonathan Hayase, and Siddhartha Srinivasa.
\newblock Git {{Re-Basin}}: {{Merging Models}} modulo {{Permutation Symmetries}}, September 2022.

\bibitem[Andreis et~al.(2023)Andreis, Bedionita, and Hwang]{andreisSetbasedNeuralNetwork2023}
Bruno Andreis, Soro Bedionita, and Sung~Ju Hwang.
\newblock Set-based {{Neural Network Encoding}}, May 2023.

\bibitem[Andriushchenko et~al.(2023)Andriushchenko, Croce, M{\"u}ller, Hein, and Flammarion]{andriushchenko2023modern}
Maksym Andriushchenko, Francesco Croce, Maximilian M{\"u}ller, Matthias Hein, and Nicolas Flammarion.
\newblock A modern look at the relationship between sharpness and generalization.
\newblock \emph{arXiv preprint arXiv:2302.07011}, 2023.

\bibitem[Ashkenazi et~al.(2022)Ashkenazi, Rimon, Vainshtein, Levi, Richardson, Mintz, and Treister]{ashkenaziNeRNLearningNeural2022}
Maor Ashkenazi, Zohar Rimon, Ron Vainshtein, Shir Levi, Elad Richardson, Pinchas Mintz, and Eran Treister.
\newblock {{NeRN}} -- {{Learning Neural Representations}} for {{Neural Networks}}, December 2022.

\bibitem[Baity-Jesi et~al.(2018)Baity-Jesi, Sagun, Geiger, Spigler, Arous, Cammarota, LeCun, Wyart, and Biroli]{baity2018comparing}
Marco Baity-Jesi, Levent Sagun, Mario Geiger, Stefano Spigler, G{\'e}rard~Ben Arous, Chiara Cammarota, Yann LeCun, Matthieu Wyart, and Giulio Biroli.
\newblock Comparing dynamics: Deep neural networks versus glassy systems.
\newblock In \emph{International Conference on Machine Learning}, pages 314--323. PMLR, 2018.

\bibitem[Belkin et~al.(2019)Belkin, Hsu, Ma, and Mandal]{belkinReconcilingModernMachinelearning2019}
Mikhail Belkin, Daniel Hsu, Siyuan Ma, and Soumik Mandal.
\newblock Reconciling modern machine-learning practice and the classical bias--variance trade-off.
\newblock \emph{Proceedings of the National Academy of Sciences}, 116\penalty0 (32):\penalty0 15849--15854, August 2019.
\newblock ISSN 0027-8424, 1091-6490.
\newblock \doi{10.1073/pnas.1903070116}.

\bibitem[Cheng and Na(2024)]{cheng2024physics}
Xiaoran Cheng and Sen Na.
\newblock Physics-informed neural networks with trust-region sequential quadratic programming.
\newblock \emph{arXiv preprint arXiv:2409.10777}, 2024.

\bibitem[Chiang et~al.(2024)Chiang, Zheng, Sheng, Angelopoulos, Li, Li, Zhang, Zhu, Jordan, Gonzalez, et~al.]{chiang2024chatbot}
Wei-Lin Chiang, Lianmin Zheng, Ying Sheng, Anastasios~Nikolas Angelopoulos, Tianle Li, Dacheng Li, Hao Zhang, Banghua Zhu, Michael Jordan, Joseph~E Gonzalez, et~al.
\newblock Chatbot arena: An open platform for evaluating llms by human preference.
\newblock \emph{arXiv preprint arXiv:2403.04132}, 2024.

\bibitem[Croce et~al.(2020)Croce, Andriushchenko, Sehwag, Debenedetti, Flammarion, Chiang, Mittal, and Hein]{croce2020robustbench}
Francesco Croce, Maksym Andriushchenko, Vikash Sehwag, Edoardo Debenedetti, Nicolas Flammarion, Mung Chiang, Prateek Mittal, and Matthias Hein.
\newblock Robustbench: a standardized adversarial robustness benchmark.
\newblock \emph{arXiv preprint arXiv:2010.09670}, 2020.

\bibitem[Cubuk et~al.(2019)Cubuk, Zoph, Shlens, and Le]{cubukRandAugmentPracticalAutomated2019}
Ekin~D. Cubuk, Barret Zoph, Jonathon Shlens, and Quoc~V. Le.
\newblock {{RandAugment}}: {{Practical}} automated data augmentation with a reduced search space, November 2019.

\bibitem[De~Luigi et~al.(2023)De~Luigi, Cardace, Spezialetti, Ramirez, Salti, and Di~Stefano]{deluigiDeepLearningImplicit2023}
Luca De~Luigi, Adriano Cardace, Riccardo Spezialetti, Pierluigi~Zama Ramirez, Samuele Salti, and Luigi Di~Stefano.
\newblock Deep {{Learning}} on {{Implicit Neural Representations}} of {{Shapes}}, February 2023.

\bibitem[Derezinski et~al.(2020)Derezinski, Liang, and Mahoney]{derezinski2020exact}
Michal Derezinski, Feynman~T Liang, and Michael~W Mahoney.
\newblock Exact expressions for double descent and implicit regularization via surrogate random design.
\newblock \emph{Advances in neural information processing systems}, 33:\penalty0 5152--5164, 2020.

\bibitem[Dosovitskiy et~al.(2021)Dosovitskiy, Beyer, Kolesnikov, Weissenborn, Zhai, Unterthiner, Dehghani, Minderer, Heigold, Gelly, Uszkoreit, and Houlsby]{dosovitskiyImageWorth16x162021}
Alexey Dosovitskiy, Lucas Beyer, Alexander Kolesnikov, Dirk Weissenborn, Xiaohua Zhai, Thomas Unterthiner, Mostafa Dehghani, Matthias Minderer, Georg Heigold, Sylvain Gelly, Jakob Uszkoreit, and Neil Houlsby.
\newblock An {{Image}} is {{Worth}} 16x16 {{Words}}: {{Transformers}} for {{Image Recognition}} at {{Scale}}.
\newblock In \emph{International {{Conference}} on {{Learning Representations}}}, 2021.

\bibitem[Eilertsen et~al.(2020)Eilertsen, J{\"o}nsson, Ropinski, Unger, and Ynnerman]{eilertsenClassifyingClassifierDissecting2020}
Gabriel Eilertsen, Daniel J{\"o}nsson, Timo Ropinski, Jonas Unger, and Anders Ynnerman.
\newblock Classifying the classifier: Dissecting the weight space of neural networks.
\newblock In \emph{{{ECAI}} 2020}. IOS Press, February 2020.

\bibitem[Engel(2001)]{engel2001statistical}
Andreas Engel.
\newblock \emph{Statistical mechanics of learning}.
\newblock Cambridge University Press, 2001.

\bibitem[Falk et~al.(2025{\natexlab{a}})Falk, Meynent, Pfammatter, Sch{\"u}rholt, and Borth]{falk2025model}
Damian Falk, L{\'e}o Meynent, Florence Pfammatter, Konstantin Sch{\"u}rholt, and Damian Borth.
\newblock A model zoo of vision transformers.
\newblock \emph{ICLR Workshop on Neural Network Weights as a New Data Modality}, 2025{\natexlab{a}}.

\bibitem[Falk et~al.(2025{\natexlab{b}})Falk, Sch{\"u}rholt, and Borth]{falk2025impact}
Damian Falk, Konstantin Sch{\"u}rholt, and Damian Borth.
\newblock The impact of model zoo size and composition on weight space learning.
\newblock \emph{ICLR Workshop on Neural Network Weights as a New Data Modality}, 2025{\natexlab{b}}.

\bibitem[Fort and Jastrzebski(2019)]{fortLargeScaleStructure2019}
Stanislav Fort and Stanislaw Jastrzebski.
\newblock Large {{Scale Structure}} of {{Neural Network Loss Landscapes}}.
\newblock In \emph{Advances in {{Neural Information Processing Systems}} 32}, June 2019.

\bibitem[Fort et~al.(2020)Fort, Dziugaite, Paul, Kharaghani, Roy, and Ganguli]{fort2020deep}
Stanislav Fort, Gintare~Karolina Dziugaite, Mansheej Paul, Sepideh Kharaghani, Daniel~M Roy, and Surya Ganguli.
\newblock Deep learning versus kernel learning: an empirical study of loss landscape geometry and the time evolution of the neural tangent kernel.
\newblock \emph{Advances in Neural Information Processing Systems}, 33:\penalty0 5850--5861, 2020.

\bibitem[Ganaie et~al.(2022)Ganaie, Hu, Malik, Tanveer, and Suganthan]{ganaie2022ensemble}
Mudasir~A Ganaie, Minghui Hu, Ashwani~Kumar Malik, Muhammad Tanveer, and Ponnuthurai~N Suganthan.
\newblock Ensemble deep learning: A review.
\newblock \emph{Engineering Applications of Artificial Intelligence}, 115:\penalty0 105151, 2022.

\bibitem[Geniesse et~al.(2024)Geniesse, Chen, Xie, Shi, Yang, Morozov, Perciano, Mahoney, Maciejewski, and Weber]{geniesse2024visualizing}
Caleb Geniesse, Jiaqing Chen, Tiankai Xie, Ge~Shi, Yaoqing Yang, Dmitriy Morozov, Talita Perciano, Michael~W Mahoney, Ross Maciejewski, and Gunther~H Weber.
\newblock Visualizing loss functions as topological landscape profiles.
\newblock \emph{arXiv preprint arXiv:2411.12136}, 2024.

\bibitem[Gokaslan and Cohen(2019)]{Gokaslan2019OpenWeb}
Aaron Gokaslan and Vanya Cohen.
\newblock Openwebtext corpus.
\newblock \url{http://Skylion007.github.io/OpenWebTextCorpus}, 2019.

\bibitem[Gordon et~al.(2021)Gordon, Duh, and Kaplan]{gordonDataParameterScaling2021}
Mitchell~A Gordon, Kevin Duh, and Jared Kaplan.
\newblock Data and {{Parameter Scaling Laws}} for {{Neural Machine Translation}}.
\newblock In \emph{Proceedings of the 2021 {{Conference}} on {{Empirical Methods}} in {{Natural Language Processing}}}, pages 5915--5922, Online and Punta Cana, Dominican Republic, November 2021. Association for Computational Linguistics.
\newblock \doi{10.18653/v1/2021.emnlp-main.478}.

\bibitem[Ha et~al.(2017)Ha, Dai, and Le]{haHyperNetworks2017}
David Ha, Andrew Dai, and Quoc~V. Le.
\newblock {{HyperNetworks}}.
\newblock In \emph{Proceedings of the {{International Conference}} on {{Learning Representations}} ({{ICLR}})}, 2017.

\bibitem[He et~al.(2016)He, Zhang, Ren, and Sun]{heDeepResidualLearning2016}
Kaiming He, Xiangyu Zhang, Shaoqing Ren, and Jian Sun.
\newblock Deep {{Residual Learning}} for {{Image Recognition}}.
\newblock In \emph{{{IEEE Conference}} on {{Computer Vision}} and {{Pattern Recognition}} ({{CVPR}})}, 2016.

\bibitem[Herrmann et~al.(2024)Herrmann, Faccio, and Schmidhuber]{herrmannLearningUsefulRepresentations2024}
Vincent Herrmann, Francesco Faccio, and J{\"u}rgen Schmidhuber.
\newblock Learning {{Useful Representations}} of {{Recurrent Neural Network Weight Matrices}}, March 2024.

\bibitem[Hestness et~al.(2017)Hestness, Narang, Ardalani, Diamos, Jun, Kianinejad, Patwary, Yang, and Zhou]{hestnessDeepLearningScaling2017}
Joel Hestness, Sharan Narang, Newsha Ardalani, Gregory Diamos, Heewoo Jun, Hassan Kianinejad, Md~Mostofa~Ali Patwary, Yang Yang, and Yanqi Zhou.
\newblock Deep {{Learning Scaling}} is {{Predictable}}, {{Empirically}}, December 2017.

\bibitem[Hoffmann et~al.(2022)Hoffmann, Borgeaud, Mensch, Buchatskaya, Cai, Rutherford, Casas, Hendricks, Welbl, Clark, Hennigan, Noland, Millican, van~den Driessche, Damoc, Guy, Osindero, Simonyan, Elsen, Rae, Vinyals, and Sifre]{hoffmannTrainingComputeOptimalLarge2022}
Jordan Hoffmann, Sebastian Borgeaud, Arthur Mensch, Elena Buchatskaya, Trevor Cai, Eliza Rutherford, Diego de~Las Casas, Lisa~Anne Hendricks, Johannes Welbl, Aidan Clark, Tom Hennigan, Eric Noland, Katie Millican, George van~den Driessche, Bogdan Damoc, Aurelia Guy, Simon Osindero, Karen Simonyan, Erich Elsen, Jack~W. Rae, Oriol Vinyals, and Laurent Sifre.
\newblock Training {{Compute-Optimal Large Language Models}}, March 2022.

\bibitem[Honegger et~al.(2023{\natexlab{a}})Honegger, Sch{\"u}rholt, and Borth]{honeggerSparsifiedModelZoo2023}
Dominik Honegger, Konstantin Sch{\"u}rholt, and Damian Borth.
\newblock Sparsified {{Model Zoo Twins}}: {{Investigating Populations}} of {{Sparsified Neural Network Models}}.
\newblock In \emph{{{ICLR}} 2023 {{Workshop}} on {{Sparsity}} in {{Neural Networks}}}. arXiv, April 2023{\natexlab{a}}.
\newblock \doi{10.48550/arXiv.2304.13718}.

\bibitem[Honegger et~al.(2023{\natexlab{b}})Honegger, Sch{\"u}rholt, Scheibenreif, and Borth]{honeggerEurosatModelZoo2023}
Dominik Honegger, Konstantin Sch{\"u}rholt, Linus Scheibenreif, and Damian Borth.
\newblock Eurosat {{Model Zoo}}: {{A Dataset}} and {{Benchmark}} on {{Populations}} of {{Neural Networks}} and {{Its Sparsified Model Twins}}.
\newblock In \emph{{{IGARSS}} 2023 - 2023 {{IEEE International Geoscience}} and {{Remote Sensing Symposium}}}, pages 888--891, July 2023{\natexlab{b}}.
\newblock \doi{10.1109/IGARSS52108.2023.10283060}.

\bibitem[Horwitz et~al.(2024)Horwitz, Shul, and Hoshen]{horwitzOriginLlamasModel2024}
Eliahu Horwitz, Asaf Shul, and Yedid Hoshen.
\newblock On the {{Origin}} of {{Llamas}}: {{Model Tree Heritage Recovery}}, May 2024.

\bibitem[HuggingFace(2025)]{huggingface}
HuggingFace.
\newblock {Hugging Face} model repository.
\newblock \url{https://huggingface.co/models}, 2025.
\newblock Accessed: 2024-05-25.

\bibitem[IARPA(2024)]{trojai}
IARPA.
\newblock Trojans in artificial intelligence (trojai).
\newblock \url{https://pages.nist.gov/trojai/}, 2024.
\newblock Accessed: 2024-05-29.

\bibitem[Ilharco et~al.(2022)Ilharco, Ribeiro, Wortsman, Gururangan, Schmidt, Hajishirzi, and Farhadi]{ilharcoEditingModelsTask2022}
Gabriel Ilharco, Marco~Tulio Ribeiro, Mitchell Wortsman, Suchin Gururangan, Ludwig Schmidt, Hannaneh Hajishirzi, and Ali Farhadi.
\newblock Editing {{Models}} with {{Task Arithmetic}}, December 2022.

\bibitem[Izmailov et~al.(2019)Izmailov, Podoprikhin, Garipov, Vetrov, and Wilson]{izmailovAveragingWeightsLeads2019a}
Pavel Izmailov, Dmitrii Podoprikhin, Timur Garipov, Dmitry Vetrov, and Andrew~Gordon Wilson.
\newblock Averaging {{Weights Leads}} to {{Wider Optima}} and {{Better Generalization}}, February 2019.

\bibitem[Jaderberg et~al.(2017)Jaderberg, Dalibard, Osindero, Czarnecki, Donahue, Razavi, Vinyals, Green, Dunning, Simonyan, et~al.]{jaderberg2017population}
Max Jaderberg, Valentin Dalibard, Simon Osindero, Wojciech~M Czarnecki, Jeff Donahue, Ali Razavi, Oriol Vinyals, Tim Green, Iain Dunning, Karen Simonyan, et~al.
\newblock Population based training of neural networks.
\newblock \emph{arXiv preprint arXiv:1711.09846}, 2017.

\bibitem[Jiang et~al.(2020)Jiang, Foret, Yak, Roy, Mobahi, Dziugaite, Bengio, Gunasekar, Guyon, and Neyshabur]{jiang2020neurips}
Yiding Jiang, Pierre Foret, Scott Yak, Daniel~M Roy, Hossein Mobahi, Gintare~Karolina Dziugaite, Samy Bengio, Suriya Gunasekar, Isabelle Guyon, and Behnam Neyshabur.
\newblock Neurips 2020 competition: Predicting generalization in deep learning.
\newblock \emph{arXiv preprint arXiv:2012.07976}, 2020.

\bibitem[Karpathy(2025)]{karpathyKarpathyNanoGPT2025}
Andrej Karpathy.
\newblock Karpathy/{{nanoGPT}}, February 2025.

\bibitem[Knyazev et~al.(2021)Knyazev, Drozdzal, Taylor, and {Romero-Soriano}]{knyazevParameterPredictionUnseen2021}
Boris Knyazev, Michal Drozdzal, Graham~W. Taylor, and Adriana {Romero-Soriano}.
\newblock Parameter {{Prediction}} for {{Unseen Deep Architectures}}.
\newblock In \emph{Conference on {{Neural Information Processing Systems}} ({{NeurIPS}})}, 2021.

\bibitem[Knyazev et~al.(2023)Knyazev, Hwang, and {Lacoste-Julien}]{knyazevCanWeScale2023}
Boris Knyazev, Doha Hwang, and Simon {Lacoste-Julien}.
\newblock Can {{We Scale Transformers}} to {{Predict Parameters}} of {{Diverse ImageNet Models}}?
\newblock In \emph{{{PMLR}}}, March 2023.

\bibitem[Kornblith et~al.(2019)Kornblith, Norouzi, Lee, and Hinton]{kornblithSimilarityNeuralNetwork2019}
Simon Kornblith, Mohammad Norouzi, Honglak Lee, and Geoffrey Hinton.
\newblock Similarity of {{Neural Network Representations Revisited}}.
\newblock In \emph{{{PMLR}}}, May 2019.

\bibitem[Krishnapriyan et~al.(2021)Krishnapriyan, Gholami, Zhe, Kirby, and Mahoney]{krishnapriyan2021characterizing}
Aditi Krishnapriyan, Amir Gholami, Shandian Zhe, Robert Kirby, and Michael~W Mahoney.
\newblock Characterizing possible failure modes in physics-informed neural networks.
\newblock \emph{Advances in Neural Information Processing Systems}, 34:\penalty0 26548--26560, 2021.

\bibitem[Krizhevsky and Hinton(2009)]{cifar}
Alex Krizhevsky and Geoffrey Hinton.
\newblock Learning multiple layers of features from tiny images.
\newblock \emph{Canadian Institute for Advanced Research}, 2009.
\newblock URL \url{http://www.cs.toronto.edu/~kriz/cifar.html}.

\bibitem[Kwon et~al.(2021)Kwon, Kim, Park, and Choi]{kwon2021asam}
Jungmin Kwon, Jeongseop Kim, Hyunseo Park, and In~Kwon Choi.
\newblock Asam: Adaptive sharpness-aware minimization for scale-invariant learning of deep neural networks.
\newblock In \emph{International Conference on Machine Learning}, pages 5905--5914. PMLR, 2021.

\bibitem[Langosco et~al.(2023)Langosco, Alex, Baker, Quarel, Bradley, and Krueger]{langosco2023detecting}
Lauro Langosco, Neel Alex, William Baker, David Quarel, Herbie Bradley, and David Krueger.
\newblock Detecting backdoors with meta-models.
\newblock In \emph{NeurIPS 2023 Workshop on Backdoors in Deep Learning-The Good, the Bad, and the Ugly}, 2023.

\bibitem[Le and Yang(2015)]{tinyimagenet}
Ya~Le and Xuan Yang.
\newblock Tiny imagenet visual recognition challenge.
\newblock \emph{CS 231N}, 7\penalty0 (7):\penalty0 3, 2015.

\bibitem[Lewkowycz et~al.(2020)Lewkowycz, Bahri, Dyer, Sohl-Dickstein, and Gur-Ari]{lewkowycz2020large}
Aitor Lewkowycz, Yasaman Bahri, Ethan Dyer, Jascha Sohl-Dickstein, and Guy Gur-Ari.
\newblock The large learning rate phase of deep learning: the catapult mechanism.
\newblock \emph{arXiv preprint arXiv:2003.02218}, 2020.

\bibitem[Li et~al.(2020)Li, Jamieson, Rostamizadeh, Gonina, Ben-Tzur, Hardt, Recht, and Talwalkar]{li2020system}
Liam Li, Kevin Jamieson, Afshin Rostamizadeh, Ekaterina Gonina, Jonathan Ben-Tzur, Moritz Hardt, Benjamin Recht, and Ameet Talwalkar.
\newblock A system for massively parallel hyperparameter tuning.
\newblock \emph{Proceedings of Machine Learning and Systems}, 2:\penalty0 230--246, 2020.

\bibitem[Liao et~al.(2020)Liao, Couillet, and Mahoney]{liao2020random}
Zhenyu Liao, Romain Couillet, and Michael~W Mahoney.
\newblock A random matrix analysis of random fourier features: beyond the gaussian kernel, a precise phase transition, and the corresponding double descent.
\newblock \emph{Advances in Neural Information Processing Systems}, 33:\penalty0 13939--13950, 2020.

\bibitem[Lim et~al.(2023)Lim, Maron, Law, Lorraine, and Lucas]{limGraphMetanetworksProcessing2023}
Derek Lim, Haggai Maron, Marc~T. Law, Jonathan Lorraine, and James Lucas.
\newblock Graph {{Metanetworks}} for {{Processing Diverse Neural Architectures}}, December 2023.

\bibitem[Liu et~al.(2019)Liu, Peng, and Schwing]{liuKnowledgeFlowImprove2019}
Iou-Jen Liu, Jian Peng, and Alexander~G. Schwing.
\newblock Knowledge {{Flow}}: {{Improve Upon Your Teachers}}.
\newblock In \emph{International {{Conference}} on {{Learning Representations}} ({{ICLR}})}, April 2019.

\bibitem[Liu et~al.(2024)Liu, Hu, Pang, Zhou, Ren, and Yang]{liu2024model}
Zihang Liu, Yuanzhe Hu, Tianyu Pang, Yefan Zhou, Pu~Ren, and Yaoqing Yang.
\newblock Model balancing helps low-data training and fine-tuning.
\newblock \emph{arXiv preprint arXiv:2410.12178}, 2024.

\bibitem[Lyu et~al.(2021)Lyu, Zhang, Sulem, and Roth]{lyu2021zero}
Qing Lyu, Hongming Zhang, Elior Sulem, and Dan Roth.
\newblock Zero-shot event extraction via transfer learning: Challenges and insights.
\newblock In \emph{Proceedings of the 59th Annual Meeting of the Association for Computational Linguistics and the 11th International Joint Conference on Natural Language Processing (Volume 2: Short Papers)}, pages 322--332, 2021.

\bibitem[Martin and Mahoney(2019{\natexlab{a}})]{martinRethinkingGeneralizationRequires2019}
Charles~H. Martin and Michael~W. Mahoney.
\newblock Rethinking generalization requires revisiting old ideas: Statistical mechanics approaches and complex learning behavior, February 2019{\natexlab{a}}.

\bibitem[Martin and Mahoney(2019{\natexlab{b}})]{martinTraditionalHeavyTailedSelf2019}
Charles~H. Martin and Michael~W. Mahoney.
\newblock Traditional and {{Heavy-Tailed Self Regularization}} in {{Neural Network Models}}.
\newblock In \emph{{{PMLR}}}, January 2019{\natexlab{b}}.

\bibitem[Martin et~al.(2021)Martin, Peng, and Mahoney]{martinPredictingTrendsQuality2021}
Charles~H. Martin, Tongsu~(Serena) Peng, and Michael~W. Mahoney.
\newblock Predicting trends in the quality of state-of-the-art neural networks without access to training or testing data.
\newblock \emph{Nature Communications}, 12\penalty0 (1):\penalty0 4122, July 2021.
\newblock ISSN 2041-1723.
\newblock \doi{10.1038/s41467-021-24025-8}.

\bibitem[Meng et~al.(2023)Meng, Sharma, Andonian, Belinkov, and Bau]{meng2023mass}
Kevin Meng, Arnab~Sen Sharma, Alex~J Andonian, Yonatan Belinkov, and David Bau.
\newblock Mass-editing memory in a transformer.
\newblock In \emph{ICLR}, 2023.

\bibitem[Mensink et~al.(2021)Mensink, Uijlings, Kuznetsova, Gygli, and Ferrari]{mensinkFactorsInfluenceTransfer2021}
Thomas Mensink, Jasper Uijlings, Alina Kuznetsova, Michael Gygli, and Vittorio Ferrari.
\newblock Factors of {{Influence}} for {{Transfer Learning}} across {{Diverse Appearance Domains}} and {{Task Types}}.
\newblock \emph{IEEE Transactions on Pattern Analysis and Machine Intelligence}, November 2021.

\bibitem[Meynent et~al.(2025)Meynent, Melev, Sch{\"u}rholt, Kauermann, and Borth]{meynent2025structure}
L{\'e}o Meynent, Ivan Melev, Konstantin Sch{\"u}rholt, G{\"o}ran Kauermann, and Damian Borth.
\newblock Structure is not enough: Leveraging behavior for neural network weight reconstruction.
\newblock \emph{ICLR Workshop on Neural Network Weights as a New Data Modality}, 2025.

\bibitem[Mitchell et~al.(2022)Mitchell, Lin, Bosselut, Finn, and Manning]{mitchellFastModelEditing2022}
Eric Mitchell, Charles Lin, Antoine Bosselut, Chelsea Finn, and Christopher~D. Manning.
\newblock Fast {{Model Editing}} at {{Scale}}.
\newblock In \emph{International {{Conference}} on {{Learning Representations}} ({{ICLR}})}. arXiv, June 2022.
\newblock \doi{10.48550/arXiv.2110.11309}.

\bibitem[Mohammed and Kora(2023)]{mohammed2023comprehensive}
Ammar Mohammed and Rania Kora.
\newblock A comprehensive review on ensemble deep learning: Opportunities and challenges.
\newblock \emph{Journal of King Saud University-Computer and Information Sciences}, 35\penalty0 (2):\penalty0 757--774, 2023.

\bibitem[Nakkiran et~al.(2019)Nakkiran, Kaplun, Bansal, Yang, Barak, and Sutskever]{nakkiranDeepDoubleDescent2019}
Preetum Nakkiran, Gal Kaplun, Yamini Bansal, Tristan Yang, Boaz Barak, and Ilya Sutskever.
\newblock Deep {{Double Descent}}: {{Where Bigger Models}} and {{More Data Hurt}}, December 2019.

\bibitem[Navon et~al.(2023)Navon, Shamsian, Achituve, Fetaya, Chechik, and Maron]{navonEquivariantArchitecturesLearning2023}
Aviv Navon, Aviv Shamsian, Idan Achituve, Ethan Fetaya, Gal Chechik, and Haggai Maron.
\newblock Equivariant {{Architectures}} for {{Learning}} in {{Deep Weight Spaces}}, January 2023.

\bibitem[Navon et~al.(2024)Navon, Shamsian, Fetaya, Chechik, Dym, and Maron]{navonEquivariantDeepWeight2024}
Aviv Navon, Aviv Shamsian, Ethan Fetaya, Gal Chechik, Nadav Dym, and Haggai Maron.
\newblock Equivariant {{Deep Weight Space Alignment}}, May 2024.

\bibitem[Netzer et~al.(2011)Netzer, Wang, Coates, Bissacco, Wu, and Ng]{svhn}
Yuval Netzer, Tao Wang, Adam Coates, Alessandro Bissacco, Bo~Wu, and Andrew~Y Ng.
\newblock Reading digits in natural images with unsupervised feature learning.
\newblock 2011.

\bibitem[OpenAI(2024)]{openai-compute}
OpenAI.
\newblock Ai and compute.
\newblock \url{https://openai.com/index/ai-and-compute/}, 2024.
\newblock Accessed: 2024-06-01.

\bibitem[Ouyang et~al.(2022)Ouyang, Beuttenmueller, G{\'o}mez-de Mariscal, Pape, Burke, Garcia-L{\'o}pez-de Haro, Russell, Moya-Sans, De-La-Torre-Guti{\'e}rrez, Schmidt, et~al.]{ouyang2022bioimage}
Wei Ouyang, Fynn Beuttenmueller, Estibaliz G{\'o}mez-de Mariscal, Constantin Pape, Tom Burke, Carlos Garcia-L{\'o}pez-de Haro, Craig Russell, Luc{\'\i}a Moya-Sans, Cristina De-La-Torre-Guti{\'e}rrez, Deborah Schmidt, et~al.
\newblock Bioimage model zoo: a community-driven resource for accessible deep learning in bioimage analysis.
\newblock \emph{BioRxiv}, pages 2022--06, 2022.

\bibitem[Peebles et~al.(2022)Peebles, Radosavovic, Brooks, Efros, and Malik]{peeblesLearningLearnGenerative2022}
William Peebles, Ilija Radosavovic, Tim Brooks, Alexei~A. Efros, and Jitendra Malik.
\newblock Learning to {{Learn}} with {{Generative Models}} of {{Neural Network Checkpoints}}, September 2022.

\bibitem[Polikar(2012)]{polikar2012ensemble}
Robi Polikar.
\newblock Ensemble learning.
\newblock \emph{Ensemble machine learning: Methods and applications}, pages 1--34, 2012.

\bibitem[Prottasha et~al.(2022)Prottasha, Sami, Kowsher, Murad, Bairagi, Masud, and Baz]{prottasha2022transfer}
Nusrat~Jahan Prottasha, Abdullah~As Sami, Md~Kowsher, Saydul~Akbar Murad, Anupam~Kumar Bairagi, Mehedi Masud, and Mohammed Baz.
\newblock Transfer learning for sentiment analysis using bert based supervised fine-tuning.
\newblock \emph{Sensors}, 22\penalty0 (11):\penalty0 4157, 2022.

\bibitem[Putterman et~al.(2024)Putterman, Lim, Gelberg, Jegelka, and Maron]{putterman2024learninglorasglequivariantprocessing}
Theo Putterman, Derek Lim, Yoav Gelberg, Stefanie Jegelka, and Haggai Maron.
\newblock Learning on loras: Gl-equivariant processing of low-rank weight spaces for large finetuned models, 2024.
\newblock URL \url{https://arxiv.org/abs/2410.04207}.

\bibitem[Radford et~al.(2019)Radford, Wu, Child, Luan, Amodei, and Sutskever]{radfordLanguageModelsAre2019}
Alec Radford, Jeffrey Wu, Rewon Child, David Luan, Dario Amodei, and Ilya Sutskever.
\newblock Language models are unsupervised multitask learners.
\newblock \emph{OpenAI blog}, 1\penalty0 (8):\penalty0 9, 2019.

\bibitem[Raissi et~al.(2019)Raissi, Perdikaris, and Karniadakis]{raissiPhysicsinformedNeuralNetworks2019}
Maziar Raissi, Paris Perdikaris, and George~E. Karniadakis.
\newblock Physics-informed neural networks: {{A}} deep learning framework for solving forward and inverse problems involving nonlinear partial differential equations.
\newblock \emph{Journal of Computational physics}, 378:\penalty0 686--707, 2019.

\bibitem[Rame et~al.(2023)Rame, Ahuja, Zhang, Cord, Bottou, and {Lopez-Paz}]{rameModelRatatouilleRecycling2023}
Alexandre Rame, Kartik Ahuja, Jianyu Zhang, Matthieu Cord, Leon Bottou, and David {Lopez-Paz}.
\newblock Model {{Ratatouille}}: {{Recycling Diverse Models}} for {{Out-of-Distribution Generalization}}.
\newblock In \emph{Proceedings of the 40th {{International Conference}} on {{Machine Learning}}}, pages 28656--28679. PMLR, July 2023.

\bibitem[Ram{\'e} et~al.(2024)Ram{\'e}, Vieillard, Hussenot, Dadashi, Cideron, Bachem, and Ferret]{rameWARMBenefitsWeight2024}
Alexandre Ram{\'e}, Nino Vieillard, L{\'e}onard Hussenot, Robert Dadashi, Geoffrey Cideron, Olivier Bachem, and Johan Ferret.
\newblock {{WARM}}: {{On}} the {{Benefits}} of {{Weight Averaged Reward Models}}, January 2024.

\bibitem[Rosenfeld et~al.(2020)Rosenfeld, Rosenfeld, Belinkov, and Shavit]{rosenfeldConstructivePredictionGeneralization2020}
Jonathan~S. Rosenfeld, Amir Rosenfeld, Yonatan Belinkov, and Nir Shavit.
\newblock A {{Constructive Prediction}} of the {{Generalization Error Across Scales}}.
\newblock In \emph{International {{Conference}} on {{Learning Representation}} ({{ICLR}})}. arXiv, 2020.
\newblock \doi{10.48550/arXiv.1909.12673}.

\bibitem[Schaeffer et~al.(2024)Schaeffer, Miranda, and Koyejo]{schaeffer2024emergent}
Rylan Schaeffer, Brando Miranda, and Sanmi Koyejo.
\newblock Are emergent abilities of large language models a mirage?
\newblock \emph{Advances in Neural Information Processing Systems}, 36, 2024.

\bibitem[Sch{\"u}rholt et~al.(2021)Sch{\"u}rholt, Kostadinov, and Borth]{schurholtSelfSupervisedRepresentationLearning2021}
Konstantin Sch{\"u}rholt, Dimche Kostadinov, and Damian Borth.
\newblock Self-{{Supervised Representation Learning}} on {{Neural Network Weights}} for {{Model Characteristic Prediction}}.
\newblock In \emph{Conference on {{Neural Information Processing Systems}} ({{NeurIPS}})}, volume~35, 2021.

\bibitem[Sch{\"u}rholt et~al.(2022{\natexlab{a}})Sch{\"u}rholt, Knyazev, {Gir{\'o}-i-Nieto}, and Borth]{schurholtHyperRepresentationsGenerativeModels2022}
Konstantin Sch{\"u}rholt, Boris Knyazev, Xavier {Gir{\'o}-i-Nieto}, and Damian Borth.
\newblock Hyper-{{Representations}} as {{Generative Models}}: {{Sampling Unseen Neural Network Weights}}.
\newblock In \emph{Thirty-Sixth {{Conference}} on {{Neural Information Processing Systems}} ({{NeurIPS}})}, September 2022{\natexlab{a}}.

\bibitem[Sch{\"u}rholt et~al.(2022{\natexlab{b}})Sch{\"u}rholt, Taskiran, Knyazev, {Gir{\'o}-i-Nieto}, and Borth]{schurholtModelZoosDataset2022}
Konstantin Sch{\"u}rholt, Diyar Taskiran, Boris Knyazev, Xavier {Gir{\'o}-i-Nieto}, and Damian Borth.
\newblock Model {{Zoos}}: {{A Dataset}} of {{Diverse Populations}} of {{Neural Network Models}}.
\newblock In \emph{Thirty-Sixth {{Conference}} on {{Neural Information Processing Systems}} ({{NeurIPS}}) {{Datasets}} and {{Benchmarks Track}}}, September 2022{\natexlab{b}}.

\bibitem[Sch{\"u}rholt et~al.(2024)Sch{\"u}rholt, Mahoney, and Borth]{schurholtScalableVersatileWeight2024}
Konstantin Sch{\"u}rholt, Michael~W. Mahoney, and Damian Borth.
\newblock Towards {{Scalable}} and {{Versatile Weight Space Learning}}.
\newblock In \emph{Proceedings of the 41st {{International Conference}} on {{Machine Learning}}}, pages 43947--43966. PMLR, July 2024.

\bibitem[Schwarze et~al.(1992)Schwarze, Opper, and Kinzel]{schwarze1992generalization}
Holm Schwarze, Manfred Opper, and Wolfgang Kinzel.
\newblock Generalization in a two-layer neural network.
\newblock \emph{Physical Review A}, 46\penalty0 (10):\penalty0 R6185, 1992.

\bibitem[Seung et~al.(1992)Seung, Sompolinsky, and Tishby]{seung1992statistical}
Hyunjune~Sebastian Seung, Haim Sompolinsky, and Naftali Tishby.
\newblock Statistical mechanics of learning from examples.
\newblock \emph{Physical review A}, 45\penalty0 (8):\penalty0 6056, 1992.

\bibitem[Shamsian et~al.(2024)Shamsian, Navon, Zhang, Zhang, Fetaya, Chechik, and Maron]{shamsianImprovedGeneralizationWeight2024}
Aviv Shamsian, Aviv Navon, David~W. Zhang, Yan Zhang, Ethan Fetaya, Gal Chechik, and Haggai Maron.
\newblock Improved {{Generalization}} of {{Weight Space Networks}} via {{Augmentations}}, February 2024.

\bibitem[Shu et~al.(2021)Shu, Kou, Cao, Wang, and Long]{shuZooTuningAdaptiveTransfer2021}
Yang Shu, Zhi Kou, Zhangjie Cao, Jianmin Wang, and Mingsheng Long.
\newblock Zoo-{{Tuning}}: {{Adaptive Transfer}} from a {{Zoo}} of {{Models}}.
\newblock In \emph{International {{Conference}} on {{Machine Learning}} ({{ICML}})}, page~12, 2021.

\bibitem[Soro et~al.(2025)Soro, Andreis, Lee, Jeong, Chong, Hutter, and Hwang]{soro2025diffusionbased}
Bedionita Soro, Bruno Andreis, Hayeon Lee, Wonyong Jeong, Song Chong, Frank Hutter, and Sung~Ju Hwang.
\newblock Diffusion-based neural network weights generation.
\newblock In \emph{The Thirteenth International Conference on Learning Representations}, 2025.
\newblock URL \url{https://openreview.net/forum?id=j8WHjM9aMm}.

\bibitem[Sorscher et~al.(2022)Sorscher, Geirhos, Shekhar, Ganguli, and Morcos]{sorscherNeuralScalingLaws2022}
Ben Sorscher, Robert Geirhos, Shashank Shekhar, Surya Ganguli, and Ari~S. Morcos.
\newblock Beyond neural scaling laws: Beating power law scaling via data pruning, November 2022.

\bibitem[Theisen et~al.(2023)Theisen, Kim, Yang, Hodgkinson, and Mahoney]{theisenWhenAreEnsembles2023}
Ryan Theisen, Hyunsuk Kim, Yaoqing Yang, Liam Hodgkinson, and Michael~W. Mahoney.
\newblock When are ensembles really effective?, May 2023.

\bibitem[Unterthiner et~al.(2020)Unterthiner, Keysers, Gelly, Bousquet, and Tolstikhin]{unterthinerPredictingNeuralNetwork2020}
Thomas Unterthiner, Daniel Keysers, Sylvain Gelly, Olivier Bousquet, and Ilya Tolstikhin.
\newblock Predicting {{Neural Network Accuracy}} from {{Weights}}.
\newblock \emph{arXiv:2002.11448 [cs, stat]}, February 2020.

\bibitem[Wang et~al.(2024)Wang, Tang, Zhao, and You]{wang2024recurrent}
Kai Wang, Dongwen Tang, Wangbo Zhao, and Yang You.
\newblock Recurrent diffusion for large-scale parameter generation, 2024.
\newblock URL \url{https://openreview.net/forum?id=CXIiV1iU3G}.

\bibitem[Wei et~al.(2022)Wei, Tay, Bommasani, Raffel, Zoph, Borgeaud, Yogatama, Bosma, Zhou, Metzler, et~al.]{wei2022emergent}
Jason Wei, Yi~Tay, Rishi Bommasani, Colin Raffel, Barret Zoph, Sebastian Borgeaud, Dani Yogatama, Maarten Bosma, Denny Zhou, Donald Metzler, et~al.
\newblock Emergent abilities of large language models.
\newblock \emph{arXiv preprint arXiv:2206.07682}, 2022.

\bibitem[Wortsman et~al.(2022{\natexlab{a}})Wortsman, Ilharco, Gadre, Roelofs, {Gontijo-Lopes}, Morcos, Namkoong, Farhadi, Carmon, Kornblith, and Schmidt]{wortsmanModelSoupsAveraging2022}
Mitchell Wortsman, Gabriel Ilharco, Samir~Yitzhak Gadre, Rebecca Roelofs, Raphael {Gontijo-Lopes}, Ari~S Morcos, Hongseok Namkoong, Ali Farhadi, Yair Carmon, Simon Kornblith, and Ludwig Schmidt.
\newblock Model soups: Averaging weights of multiple fine-tuned models improves accuracy without increasing inference time.
\newblock In \emph{International {{Conference}} on {{Machine Learning}}}, 2022{\natexlab{a}}.

\bibitem[Wortsman et~al.(2022{\natexlab{b}})Wortsman, Ilharco, Kim, Li, Kornblith, Roelofs, {Gontijo-Lopes}, Hajishirzi, Farhadi, Namkoong, and Schmidt]{wortsmanRobustFinetuningZeroshot2022}
Mitchell Wortsman, Gabriel Ilharco, Jong~Wook Kim, Mike Li, Simon Kornblith, Rebecca Roelofs, Raphael {Gontijo-Lopes}, Hannaneh Hajishirzi, Ali Farhadi, Hongseok Namkoong, and Ludwig Schmidt.
\newblock Robust fine-tuning of zero-shot models.
\newblock In \emph{Conference on {{Computer Vision}} and {{Pattern Recognition}} ({{CVPR}})}, 2022{\natexlab{b}}.

\bibitem[Yak et~al.(2018)Yak, Gonzalvo, and Mazzawi]{yakTaskArchitectureIndependentGeneralization2018}
Scott Yak, Javier Gonzalvo, and Hanna Mazzawi.
\newblock Towards {{Task}} and {{Architecture-Independent Generalization Gap Predictors}}.
\newblock In \emph{Advances in Neural Information Processing Systems}, volume~31, 2018.

\bibitem[Yang et~al.(2021)Yang, Hodgkinson, Theisen, Zou, Gonzalez, Ramchandran, and Mahoney]{yangTaxonomizingLocalGlobal2021}
Yaoqing Yang, Liam Hodgkinson, Ryan Theisen, Joe Zou, Joseph~E Gonzalez, Kannan Ramchandran, and Michael~W Mahoney.
\newblock Taxonomizing local versus global structure in neural network loss landscapes.
\newblock In \emph{Advances in {{Neural Information Processing Systems}}}, volume~34, pages 18722--18733. Curran Associates, Inc., 2021.

\bibitem[Yao et~al.(2018)Yao, Gholami, Lei, Keutzer, and Mahoney]{yao2018hessian}
Zhewei Yao, Amir Gholami, Qi~Lei, Kurt Keutzer, and Michael~W Mahoney.
\newblock Hessian-based analysis of large batch training and robustness to adversaries.
\newblock \emph{Advances in Neural Information Processing Systems}, 31, 2018.

\bibitem[Yosinski et~al.(2014)Yosinski, Clune, Bengio, and Lipson]{yosinskiHowTransferableAre2014}
Jason Yosinski, Jeff Clune, Yoshua Bengio, and Hod Lipson.
\newblock How transferable are features in deep neural networks?
\newblock In \emph{Neural {{Information Processing Systems}} ({{NeurIPS}})}, November 2014.

\bibitem[You(2018)]{pytorchcv2018}
Donny You.
\newblock \url{https://github.com/donnyyou/PyTorchCV}, 2018.

\bibitem[Zdeborov{\'a} and Krzakala(2016)]{zdeborova2016statistical}
Lenka Zdeborov{\'a} and Florent Krzakala.
\newblock Statistical physics of inference: Thresholds and algorithms.
\newblock \emph{Advances in Physics}, 65\penalty0 (5):\penalty0 453--552, 2016.

\bibitem[Zhang et~al.(2019)Zhang, Ren, and Urtasun]{zhangGraphHyperNetworksNeural2019}
Chris Zhang, Mengye Ren, and Raquel Urtasun.
\newblock Graph {{HyperNetworks}} for {{Neural Architecture Search}}.
\newblock In \emph{International {{Conference}} on {{Learning Representations}} ({{ICLR}})}, 2019.

\bibitem[Zhang et~al.(2023)Zhang, Kofinas, Zhang, Chen, Burghouts, and Snoek]{zhangNeuralNetworksAre2023}
David~W Zhang, Miltiadis Kofinas, Yan Zhang, Yunlu Chen, Gertjan~J Burghouts, and Cees G~M Snoek.
\newblock Neural {{Networks Are Graphs}}!{{Graph Neural Networks}} for {{Equivariant Processing}} of {{Neural Networks}}.
\newblock July 2023.

\bibitem[Zheng et~al.(2024)Zheng, Chiang, Sheng, Zhuang, Wu, Zhuang, Lin, Li, Li, Xing, et~al.]{zheng2024judging}
Lianmin Zheng, Wei-Lin Chiang, Ying Sheng, Siyuan Zhuang, Zhanghao Wu, Yonghao Zhuang, Zi~Lin, Zhuohan Li, Dacheng Li, Eric Xing, et~al.
\newblock Judging llm-as-a-judge with mt-bench and chatbot arena.
\newblock \emph{Advances in Neural Information Processing Systems}, 36, 2024.

\bibitem[Zhou et~al.(2023{\natexlab{a}})Zhou, Yang, Burns, Jiang, Sokota, Kolter, and Finn]{zhouPermutationEquivariantNeural2023}
Allan Zhou, Kaien Yang, Kaylee Burns, Yiding Jiang, Samuel Sokota, J.~Zico Kolter, and Chelsea Finn.
\newblock Permutation {{Equivariant Neural Functionals}}, February 2023{\natexlab{a}}.

\bibitem[Zhou et~al.(2023{\natexlab{b}})Zhou, Yang, Chang, and Mahoney]{zhou2023three}
Yefan Zhou, Yaoqing Yang, Arin Chang, and Michael~W Mahoney.
\newblock A three-regime model of network pruning.
\newblock In \emph{International Conference on Machine Learning}, pages 42790--42809. PMLR, 2023{\natexlab{b}}.

\bibitem[Zhou et~al.(2024)Zhou, Chen, Cao, Schürholt, and Yang]{zhou2024MD}
Yefan Zhou, Jianlong Chen, Qinxue Cao, Konstantin Schürholt, and Yaoqing Yang.
\newblock Md tree: a model-diagnostic tree grown on loss landscape.
\newblock In \emph{International Conference on Machine Learning}, 2024.

\end{thebibliography}

\newpage
\appendix

\section{Dataset documentation}
\label{app:dataset_doc}

Please find the dataset at \url{https://phasetransitions.modelzoos.cc}\footnote{Back-up link: \url{https://github.com/ModelZoos/PhaseTransitionModelZoo}}. The repository contains instructions on how to access the dataset, code to use it, code used to generate the model zoos and the loss landscape metrics, as well as other related information.

\subsection{Model Zoo Contents}
In the main paper, we described the generation of the model zoos as well as explored their performance and phase information. 
Here, we detail the contents of the datasets.
A model zoo contains a set of trained Neural Network models. For each of the zoos, we fix architecture and task combinations and introduce variations in temperature-like and load-like parameters. We realize temperature variations by varying the batch-size, and load variations by varying the model width. 
We chose the training parameters and variation range such that the phases and phase transitions described by Yang et. al~\citep{yangTaxonomizingLocalGlobal2021} can be observed. 
We repeat each temperature-load combination 3 times with different random seeds to compute loss landscape metrics and get robust results.

For every model sample, there are model state checkpoints at intervals throughout training. The checkpoints are in PyTorch format, which uses pickle to save ordered dicts. We will provide code to convert the checkpoints to framework-neutral file formats.
We annotate these samples with performance metrics (training and test loss and accuracy), as well as the loss landscape metrics (hessian eigenvalues, Bézier mode connectivity, CKA similarity). 
We add additional results like model averaging performance, where applicable to individual models. 
The model zoos are generated with \texttt{ray.tune}~\footnote{\url{https://docs.ray.io/en/latest/tune/index.html}} and largely follow their experiment structure. Each model in a population is contained in one folder. Checkpoints are kept in subfolders for the corresponding epochs. Each model is annotated with a \texttt{config.json} file to re-create the model exactly. Performance metrics are tracked for every epoch and saved in a \texttt{results.json} file for every model. For a subset of epochs, we add loss-landscape metrics.
All model zoos contain full meta-data configs and self-contained Pytorch code, s.t. they can be re-instantiated exactly, re-trained, or fine-tuned. 
All code to train grids, evaluate, compute loss landscape metrics and model averaging is available alongside the data.
Further, we provide code to i) recreate the model zoo datasets, ii) compute loss-landscape metrics, iii) load the models, and iv) re-create the figures in the main paper. 
In order to allow easy use of the dataset, we plan to make adequate PyTorch dataset classes
available upon publication. 

This section will be updated upon dataset publication. Indeed, several statements are intentionally left vague as of now. Our dataset is large and will require a careful choice of what to include to balance the dataset's utility with its size. This will influence, in particular, the number of checkpoints that we include per model.

\subsection{Model Zoo Generation}
\label{app:model_zoo_gen}
We generated the dataset for common computer vision, language, and SciML tasks and architectures to show generalization and maximize applicability to the community. 
We fixed the load-temperature grids by exploring the boundary cases first and establishing the presence of phase transitions, then filling in more resolution.
For the vision and language zoos, we chose batch size as a temperature-like parameter, as it's a common hyperparameter to vary. For the load-like parameter, we vary the width of the models by changing the number of channels or the size of the internal token, respectively.
For the SciML zoo, we adapt the scheme to the specifics of the domain. To vary the temperature, we chose to change the learning rate to keep the batch size and collocation points constant. To vary the load, we change the complexity of the domain by varying $\beta$ parameter of the physical system.
We base the GPT-model implementation and training setup on NanoGPT~\cite{karpathyKarpathyNanoGPT2025} and use Chinchilla scaling laws to find the ideal number of training steps~\cite{hoffmannTrainingComputeOptimalLarge2022}.
The full list of model zoo hyperparameters is given in Tables \ref{tab:hyperparams_vision},\ref{tab:hyperparams_language} and \ref{tab:hyperparams_sciml}. For vision zoos, we use Random Cropping, horizontal flipping, and random rotations. Training ViTs on CIFAR100 required stronger data augmentation to achieve competitive performance. Therefore, we have applied a combination of random cropping, random erasing, color jitter, and RandAugment~\citep{cubukRandAugmentPracticalAutomated2019}.
After the initial tuning of the grids, the training of the model zoos was done on 16 DGX H100 GPUS in 30 days. The computation of loss landscape metrics was performed on the same hardware in 14 days.

\begin{table}[t]
\captionof{table}{
Summary of the content of each model zoo included in our dataset.
}
\setlength{\tabcolsep}{5pt}
\label{tab:model_zoo_summary}
\centering
\small
\begin{tabularx}{1.0\linewidth}{ccccc}
\toprule
\textbf{Architecture} & \textbf{Dataset}       & \textbf{\# models} & \textbf{Load-like param.} & \textbf{Temperature-like param.} \\
\cmidrule(lr){1-1}\cmidrule(lr){2-2}\cmidrule(lr){3-3}\cmidrule(lr){4-4}\cmidrule(lr){5-5}
ResNet-18    & SVHN          & 192      & Model width              & Batch size \\
ResNet-18    & CIFAR-10      & 192      & Model width              & Batch size \\
ResNet-18    & CIFAR-100     & 192      & Model width              & Batch size \\
ResNet-18    & TinyImagenet  & 192      & Model width              & Batch size \\
ResNet-50    & SVHN          & 192      & Model width              & Batch size \\
ResNet-50    & CIFAR-10      & 192      & Model width              & Batch size \\
ResNet-50    & CIFAR-100     & 192      & Model width              & Batch size \\
ResNet-50    & TinyImagenet  & 192      & Model width              & Batch size \\
ViT          & CIFAR-10      & 147      & Model width              & Batch size \\
ViT          & CIFAR-100     & 147      & Model width              & Batch size \\
GPT-2        & OpenWebText   & 264      & \texttt{model\_dim}      & Batch size \\
MLP (PINN)   & 1D Convection & 700      & \texttt{beta}            & Learning rate \\
\bottomrule
\end{tabularx}
\end{table}

\begin{table}[h]
\captionof{table}{Full list of hyperparameters of the vision model zoos. Width indicates the width of the first residual block. From that, we follow the same scaling factor as the standard ResNet.}
\label{tab:hyperparams_vision}
\centering
\small
\setlength{\tabcolsep}{5.0pt}
\begin{tabularx}{1.0\linewidth}{lccc}
\toprule
Base Architecture       &  & ResNet-18, ResNet-50                                                       & ViT                                                                                                                                                         \\
\multirow{2}{*}{Datasets}                &  & SVHN, CIFAR10                          & \multicolumn{1}{c}{CIFAR10, CIFAR100}                                                                                              \\
                        &  & CIFAR100, TinyImagenet                                                       & \multicolumn{1}{l}{}                                                                                                                                        \\
Activation              &  & ReLU                                                                       & ReLU                                                                                                                                                        \\
Initialization          &  & Kaiming Uniform                                                            & Kaiming Uniform                                                                                                                                             \\
Optimizer               &  & SGD                                                                        & ADAMW                                                                                                                                                       \\
Learning Rate           &  & 0.1                                                                        & $6e-3$                                                                                                                                                      \\
Momentum                &  & 0.9                                                                        & \multicolumn{1}{l}{}                                                                                                                                        \\
WD                      &  & $5e-4$                                                                     & CIFAR10: $5e-4$. CIFAR100: $5e-2$                                                                                                                           \\
LR Schedule             &  & OneCycleLR with Cosine Annealing                                           & OneCycleLR with Cosine Annealing                                                                                                                            \\
Width                   &  & 2, 4, 8, 16, 32, 64, 128, 256                                              & 6, 12, 24, 48, 60, 96, 192                                                                                                                                  \\
Batch Size              &  & 8, 16, 32, 64, 128, 256, 512, 1024                                         & 8, 16, 32, 64, 128, 256, 512                                                                                                                                \\
Seeds                   &  & 0, 1, 2                                                                    & 0, 1, 2                                                                                                                                                     \\
\bottomrule
\end{tabularx}
\end{table}

\begin{table}[h]
\captionof{table}{Full list of hyperparameters of the lanugage model zoos. \texttt{model\_dim} indicates the tokensize of the transformer blocks.}
\label{tab:hyperparams_language}
\centering
\small
\setlength{\tabcolsep}{25.0pt}
\begin{tabularx}{1.0\linewidth}{lcc}
\toprule
Base Architecture       &  & GPT-2 \\
\multirow{1}{*}{Dataset}                &  & OpenWebText  \\
Activation              &  & GeLU \\
Initialization          &  & Kaiming Uniform          \\
Optimizer               &  & ADAMW             \\
Learning Rate           &  & $6e-4$            \\
WD                      &  & $1e-1$            \\
LR Schedule             &  & OneCycleLR with Cosine Annealing                  \\
\texttt{model\_dim}                   &  & 24, 48, 96, 192, 276, 384, 540, 768, 1092, 1536, 2172 \\
Batch Size              &      &  32, 64, 128, 256, 512, 1024, 2048, 4096 \\
Seeds                   &  & 1, 2, 3                                      \\
Block size              & &   1024  \\
Training steps              & &   350k \\  
\bottomrule
\end{tabularx}
\end{table}

\begin{table}[h]
\captionof{table}{Full list of hyperparameters of the SciML model zoos. }
\label{tab:hyperparams_sciml}
\centering
\small
\setlength{\tabcolsep}{15pt}
\begin{tabularx}{1.0\linewidth}{lcc}
\toprule
Base Architecture       &  &  MLP \\
Problem  &  & 1D convection \\
Activation   &  & Tanh   \\
Initialization  &  &  Kaiming Uniform   \\
Optimizer               &  &  LBFGS      \\
Learning Rate           &  & $0.0001$, $0.0005$, $0.001$, $0.005$, $0.01$, $0.05$, $0.1$, $0.5$, $1.0$, $1.5$ \\
Convection coefficient $\beta$ & & $1$, $2$, $3$, $4$, $5$, $6$, $7$, $8$, $9$, $10$, $15$, $30$, $50$, $70$ \\
Line search function   & & Strong Wolfe \\
\texttt{model\_dim}    &  & (50,50,50,50,1) \\
Batch Size             &  &  Full batch \\
Seeds                   &  &   0, 123, 2023, 54321, 123456                                      \\
Max iteration        & &   100k \\ 
Histort size         &  & 50 \\
Loss                 & &   MSE  \\
Collocation points  & &  100 \\
\bottomrule
\end{tabularx}
\end{table}

\subsection{Model zoo evaluation}
\label{app:zoo_eval}
\begin{figure}
    \centering
    \includegraphics[width=\textwidth]{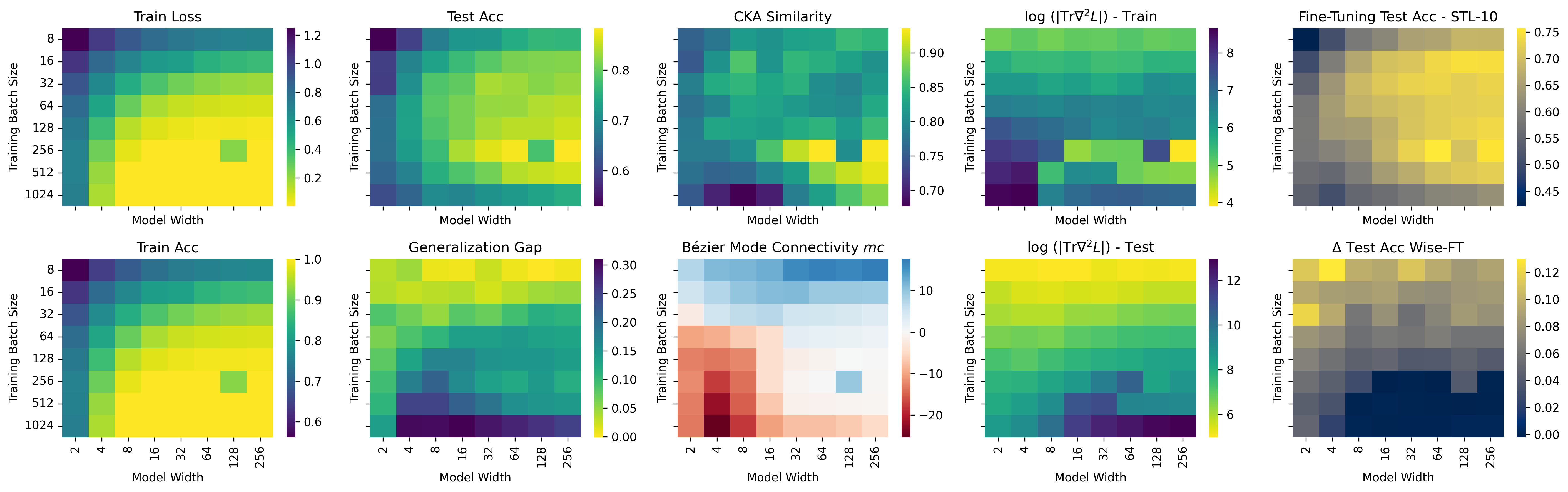}
    \caption{Phase plots for the CIFAR-10 ResNet-18 model zoo,  showing distinct phase transitions in performance and loss-landscape metrics, fine-tuninng, and weight averaging.}
    \label{fig:phase_plot_r18_c10}
\end{figure}

\begin{figure}
    \centering
    \includegraphics[width=\textwidth]{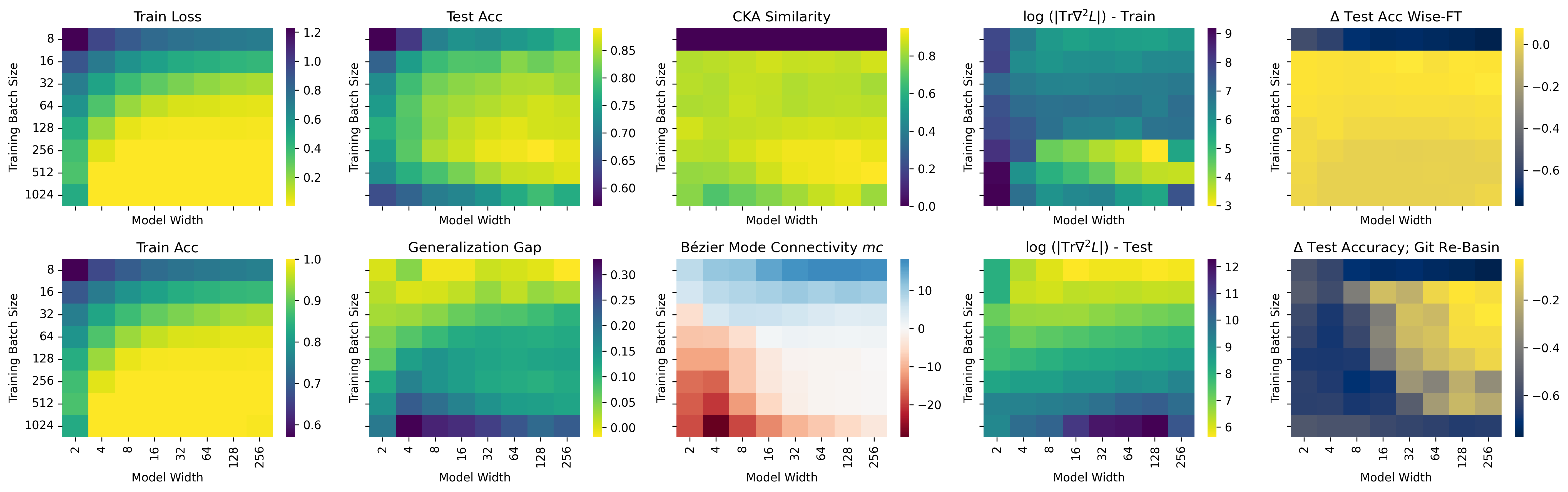}
    \caption{Phase plots for the CIFAR-10 ResNet-50 model zoo,  showing distinct phase transitions in performance and loss-landscape metrics, and weight averaging.}
    \label{fig:phase_plot_r50_c10}
\end{figure}

\begin{figure}
    \centering
    \includegraphics[width=\textwidth]{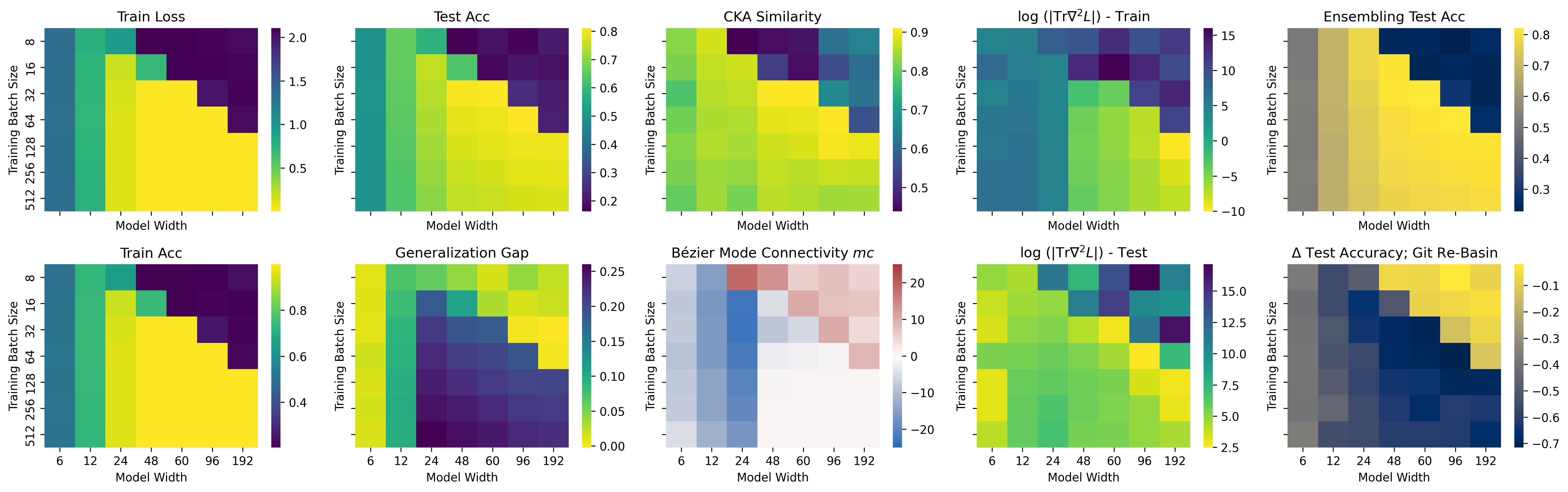}
    \caption{Phase plots for the CIFAR-10 ViT model zoo, showing distinct phase transitions in performance and loss-landscape metrics, ensembling, and weight averaging.}
    \label{fig:phase_plot_vit_c10}
\end{figure}

\begin{figure}
    \centering
    \includegraphics[width=\textwidth]{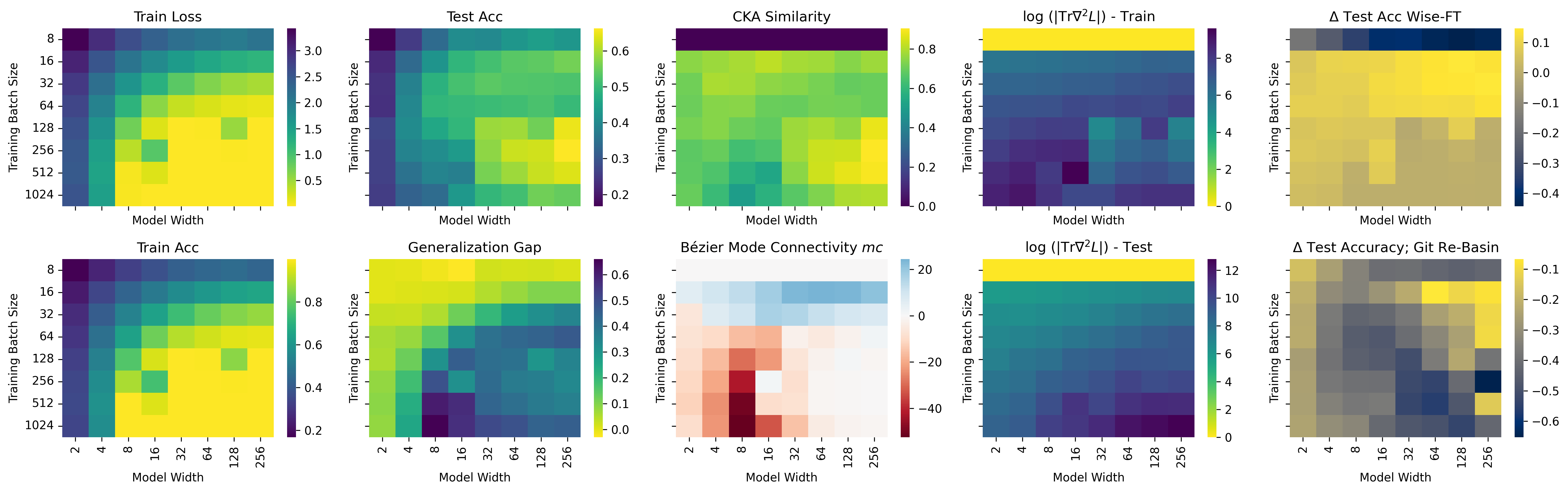}
    \caption{Phase plots for the CIFAR-100 ResNet-18 model zoo, showing distinct phase transitions in performance and loss-landscape metrics, and weight averaging.}
    \label{fig:phase_plot_r18_c100}
\end{figure}

\begin{figure}
    \centering
    \includegraphics[width=\textwidth]{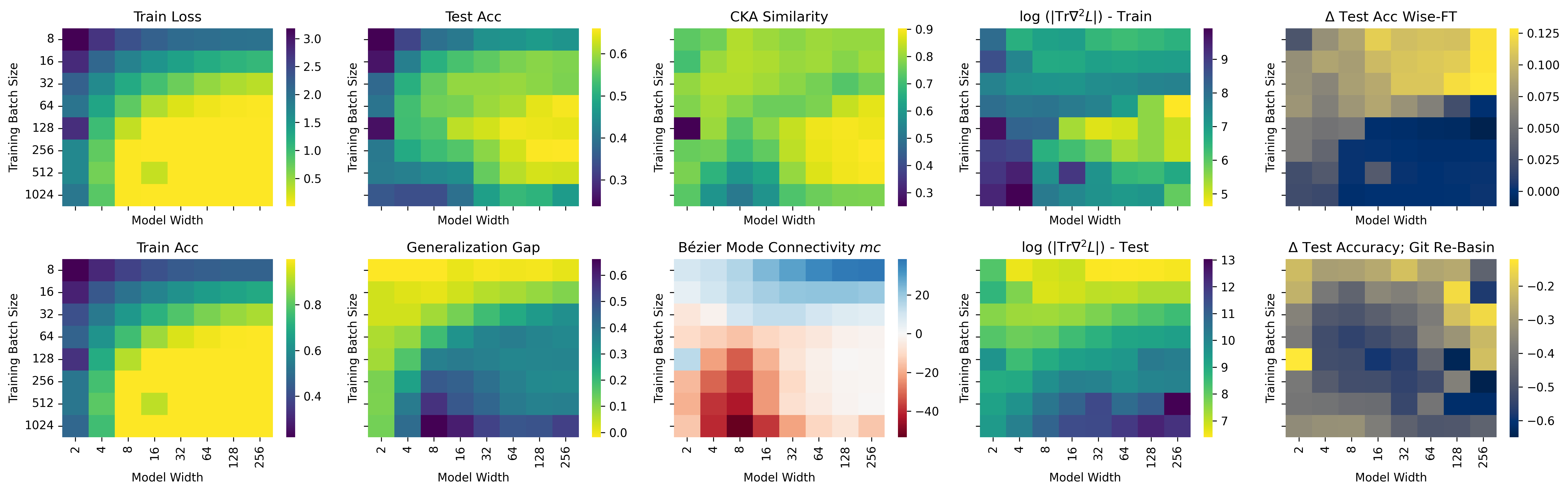}
    \caption{Phase plots for the CIFAR-100 ResNet-50 model zoo, showing distinct phase transitions performance, loss landscape metrics, and weight averaging.}
    \label{fig:phase_plot_r50_c100}
\end{figure}

\begin{figure}
    \centering
    \includegraphics[width=\textwidth]{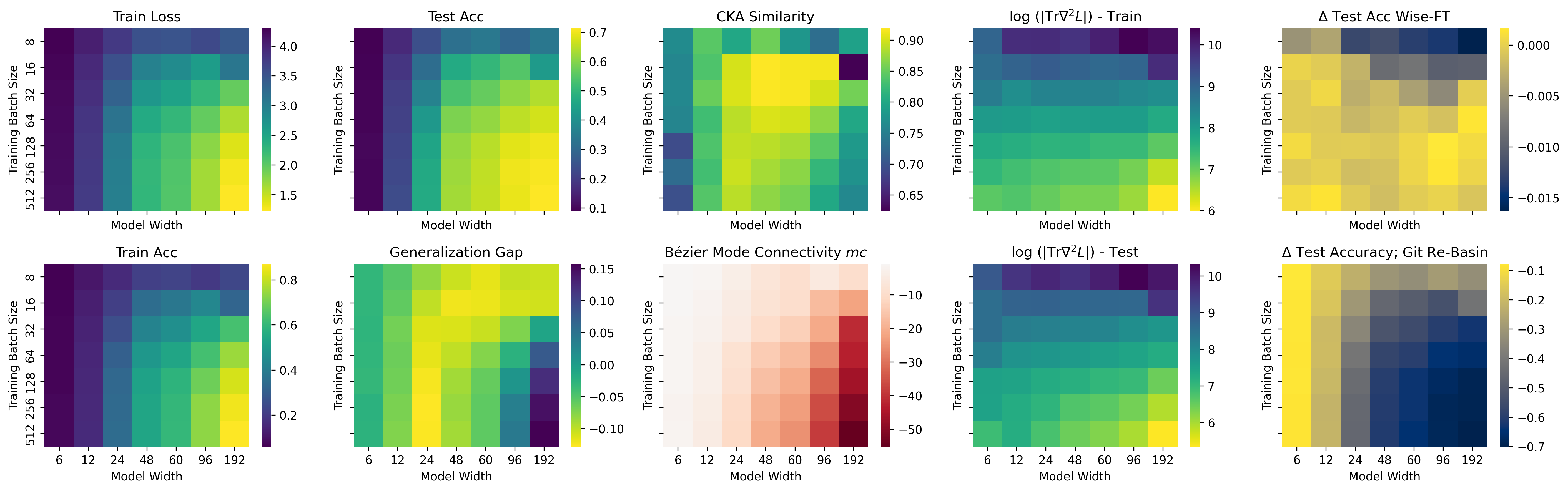}
    \caption{Phase plots for the CIFAR-100 ViT model zoo, showing distinct phase transitions in performance, loss landscape metrics, transfer learning and weight averaging.}
    \label{fig:phase_plot_vit_c100}
\end{figure}

\begin{figure}
    \centering
    \includegraphics[width=\textwidth]{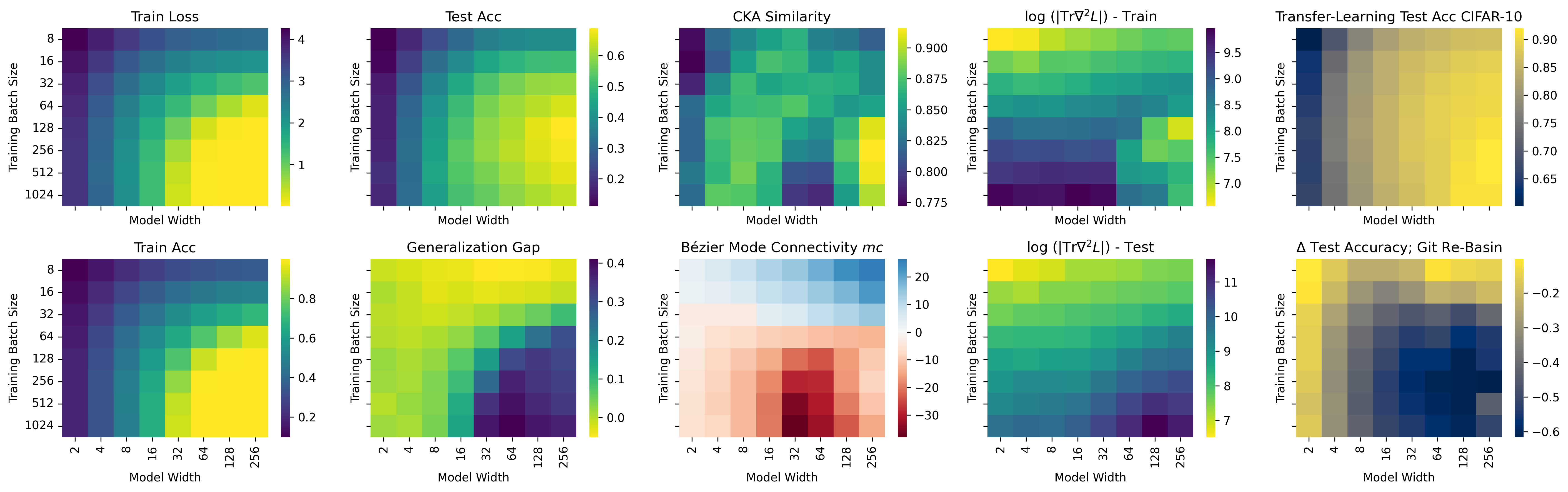}
    \caption{Phase plots for the Tiny-Imagenet ResNet-18 model zoo, showing distinct phase transitions in performance and loss-landscape metrics.}
    \label{fig:phase_plot_r18_ti}
\end{figure}

\begin{figure}
    \centering
    \includegraphics[width=\textwidth]{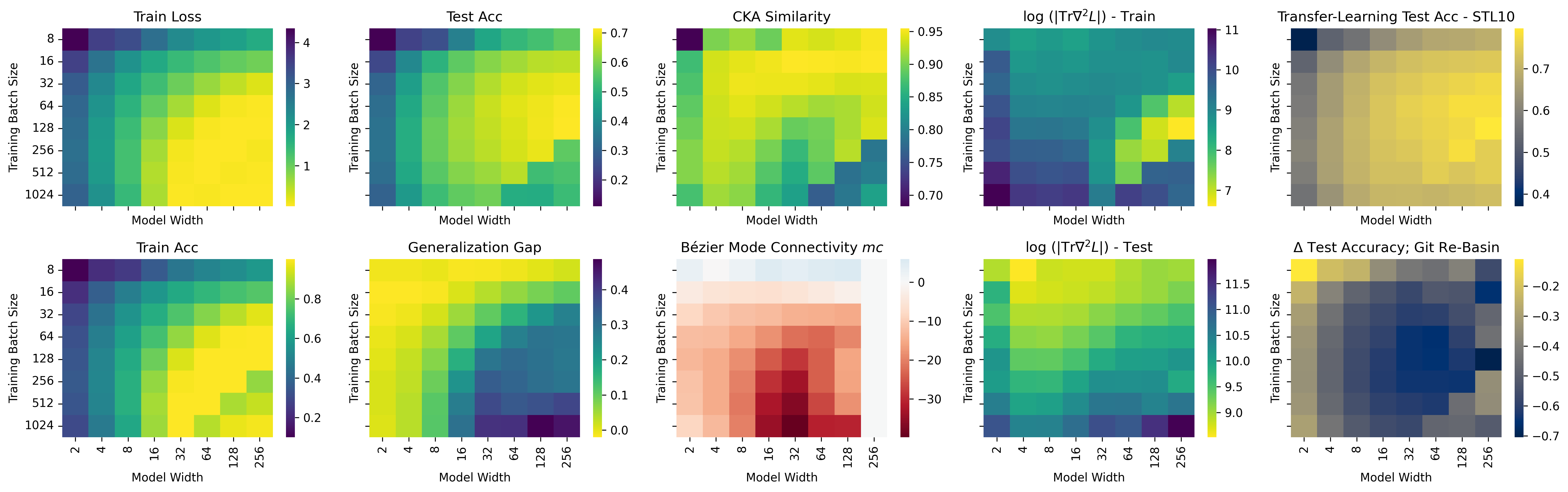}
    \caption{Phase plots for the Tiny-Imagenet ResNet-50 model zoo, showing distinct phase transitions in performance, loss landscape metrics, transfer learning and weight averaging.}
    \label{fig:phase_plot_r50_ti}
\end{figure}

\begin{figure}
    \centering
    \includegraphics[width=\textwidth]{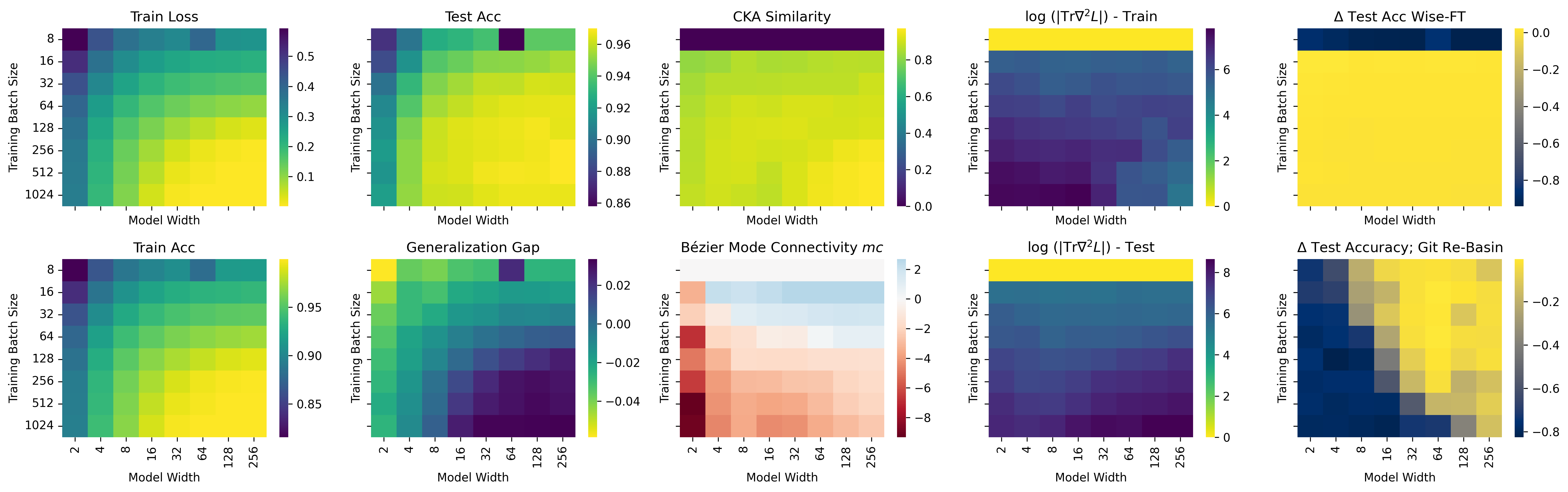}
    \caption{Phase plots for the SVHN ResNet-18 model zoo,  showing distinct phase transitions in performance and loss-landscape metrics, and weight averaging..}
    \label{fig:phase_plot_r18_svhn}
\end{figure}

\begin{figure}
    \centering
    \includegraphics[width=\textwidth]{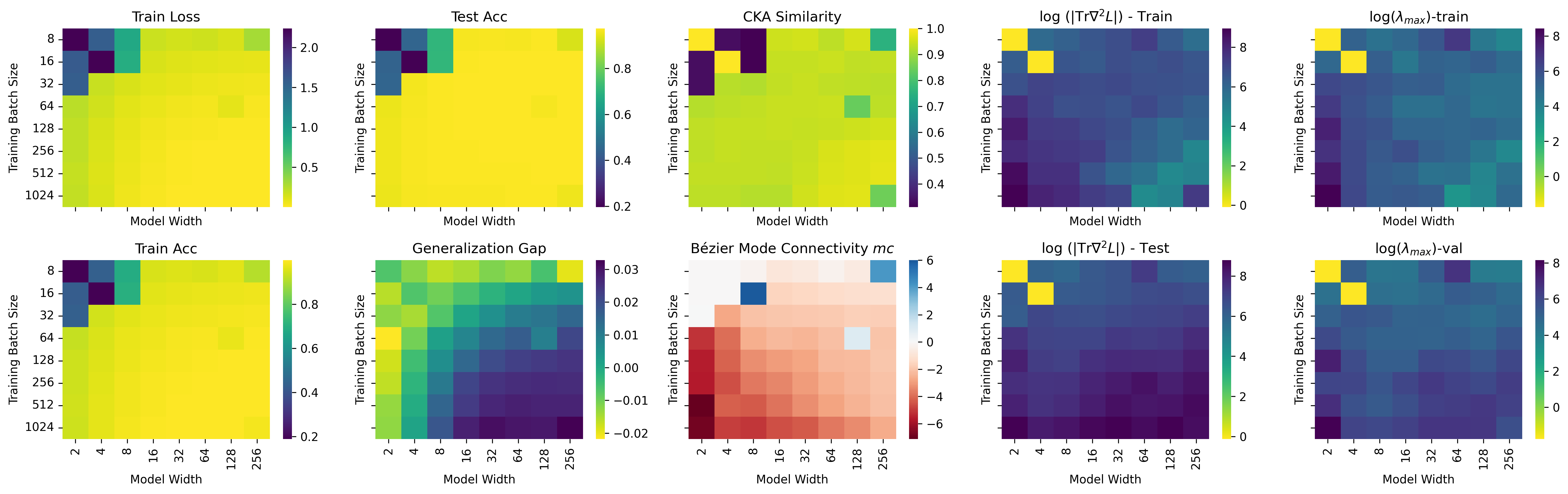}
    \caption{Phase plots for the SVHN ResNet-50 model zoo, showing distinct phase transitions in performance and loss-landscape metrics.
    }
    \label{fig:phase_plot_r50_svhn}
\end{figure}

\begin{figure}
    \centering
    \includegraphics[width=\textwidth]{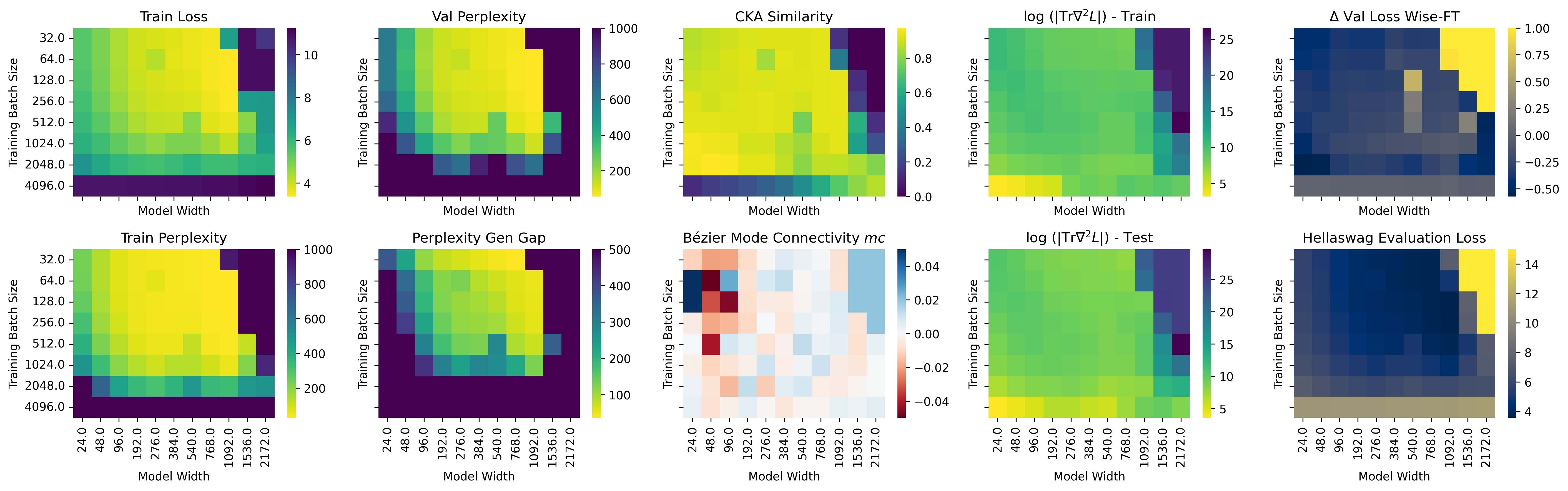}
    \caption{Phase plots for the openwebtext GPT-2 model zoo, showing distinct phase transitions in in performance and loss-landscape metrics, aligned with the vision model zoos. Please note that the reported metrics are adjusted to the language domain. Instead of accuracy, we report perplexity. Also, the bezier mode connectivity is computed on the loss, rather than on the error, which induces some noisiness.}
    \label{fig:phase_plot_gpt2_openwebtext}
\end{figure}

In this section, we test the general validity of the trained models as representatives of real-world models in a structured dataset. An overview of the models at the end of training is given in Tables \ref{tab:zoo_performance} and \ref{tab:zoo_performance_gpt2}. The results confirm that models are trained to competitive performance for their respective sizes.\looseness-1
More nuanced information on the distribution of model performance on the temperature-load grid is shown in Figures \ref{fig:phase_plot_r18_c10} through \ref{fig:phase_plot_gpt2_openwebtext}. Similar to previous work, the zoos show distinct low train-loss regions, with smaller embedded regions within that generalize well. 
Test performance generally improves with decreasing load (increasing width), with a distinct peak phase where temperature and load are low enough, but not too low. The generalization gap correspondingly shows a superposition of both patterns. Further applications or loss landscape metrics likewise show clear phase transitions.
Remarkably, the phase layout and loss landscape metrics generalize across different domains, tasks, architectures, and datasets.\looseness-1


\begin{table}[h!]
\captionof{table}{Conventional Performance Metric Distribution of the Computer Vision Model Zoos.\looseness-1}
\label{tab:zoo_performance}
\centering
\tiny
\setlength{\tabcolsep}{4.5pt}

\begin{tabularx}{1.0\linewidth}{ccccccc}
\toprule
\textbf{Arch.}       & \textbf{Data}        & {\textbf{Train Loss}}        & {\textbf{Test Loss}}         & {\textbf{Train Acc}}         & {\textbf{Test Acc}}          & {\textbf{GGap}}              \\
\multicolumn{1}{l}{} & \multicolumn{1}{l}{} & {$\mu \pm \sigma$ [min,max]} & {$\mu \pm \sigma$ [min,max]} & {$\mu \pm \sigma$ [min,max]} & {$\mu \pm \sigma$ [min,max]} & {$\mu \pm \sigma$ [min,max]} \\
\cmidrule(lr){1-1} \cmidrule(lr){2-2} \cmidrule(lr){3-3} \cmidrule(lr){4-4} \cmidrule(lr){5-5} \cmidrule(lr){6-6} \cmidrule(lr){7-7}
R-18             & SVHN                 & 0.10$\pm$0.11 {[}0.00,0.38{]} & 0.15$\pm$0.04 {[}0.11,0.27{]} & 0.97$\pm$0.03 {[}0.88,1.00{]} & 0.96$\pm$0.01 {[}0.92,0.97{]} & 0.01$\pm$0.02 {[}-0.04,0.03{]} \\
R-50             & SVHN                 & 0.06$\pm$0.07 {[}0.00,0.24{]} & 0.14$\pm$0.02 {[}0.11,0.18{]} & 0.98$\pm$0.02 {[}0.93,1.00{]} & 0.97$\pm$0.01 {[}0.95,0.97{]} & 0.01$\pm$0.02 {[}-0.02,0.03{]} \\
\cmidrule(lr){1-1} \cmidrule(lr){2-2} \cmidrule(lr){3-3} \cmidrule(lr){4-4} \cmidrule(lr){5-5} \cmidrule(lr){6-6} \cmidrule(lr){7-7}
R-18             & C-10              & 0.08$\pm$0.19 {[}0.00,0.66{]} & 0.67$\pm$0.34 {[}0.32,1.98{]} & 0.97$\pm$0.06 {[}0.77,1.00{]} & 0.82$\pm$0.08 {[}0.65,0.91{]} & 0.16$\pm$0.07 {[}0.04,0.35{]}  \\
R-50             & C-10              & 0.04$\pm$0.09 {[}0.00,0.52{]} & 0.60$\pm$0.30 {[}0.27,1.69{]} & 0.99$\pm$0.03 {[}0.82,1.00{]} & 0.84$\pm$0.07 {[}0.64,0.92{]} & 0.15$\pm$0.06 {[}0.05,0.33{]}  \\
\cmidrule(lr){1-1} \cmidrule(lr){2-2} \cmidrule(lr){3-3} \cmidrule(lr){4-4} \cmidrule(lr){5-5} \cmidrule(lr){6-6} \cmidrule(lr){7-7}
R-18             & C-100             & 0.45$\pm$0.79 {[}0.00,2.48{]} & 2.02$\pm$0.54 {[}1.24,3.89{]} & 0.88$\pm$0.21 {[}0.35,1.00{]} & 0.53$\pm$0.12 {[}0.29,0.69{]} & 0.35$\pm$0.15 {[}0.01,0.67{]}  \\
R-50             & C-100             & 0.35$\pm$0.68 {[}0.00,4.61{]} & 1.78$\pm$0.55 {[}1.18,4.61{]} & 0.91$\pm$0.18 {[}0.01,1.00{]} & 0.57$\pm$0.11 {[}0.01,0.70{]} & 0.34$\pm$0.14 {[}-0.01,0.67{]} \\
\cmidrule(lr){1-1} \cmidrule(lr){2-2} \cmidrule(lr){3-3} \cmidrule(lr){4-4} \cmidrule(lr){5-5} \cmidrule(lr){6-6} \cmidrule(lr){7-7}
R-18             & TI         & 1.20$\pm$1.06 {[}0.01,3.42{]} & 1.91$\pm$0.48 {[}1.29,3.22{]} & 0.71$\pm$0.25 {[}0.23,1.00{]} & 0.55$\pm$0.12 {[}0.26,0.70{]} & 0.16$\pm$0.15 {[}0.03,0.41{]}  \\
R-50             & TI         & 1.05$\pm$0.96 {[}0.00,3.55{]} & 1.85$\pm$0.51 {[}1.21,3.63{]} & 0.74$\pm$0.22 {[}0.21,1.00{]} & 0.57$\pm$0.11 {[}0.22,0.72{]} & 0.17$\pm$0.15 {[}-0.02,0.49{]}\\
\cmidrule(lr){1-1} \cmidrule(lr){2-2} \cmidrule(lr){3-3} \cmidrule(lr){4-4} \cmidrule(lr){5-5} \cmidrule(lr){6-6} \cmidrule(lr){7-7}
VIT                  & C-10              & 0.77$\pm$0.83 {[}0.00,2.18{]} & 1.72$\pm$0.45 {[}0.71,2.96{]} & 0.71$\pm$0.31 {[}0.17,1.00{]} & 0.59$\pm$0.23 {[}0.10,0.82{]} & 0.13$\pm$0.10 {[}-0.01,0.27{]}  \\
VIT                  & C-100             & 2.96$\pm$0.94 {[}1.21,4.32{]} & 2.72$\pm$0.82 {[}1.74,4.15{]} & 0.37$\pm$0.24 {[}0.06,0.88{]} & 0.43$\pm$0.22 {[}0.09,0.72{]} & -0.05$\pm$0.07 {[}-0.13,0.16{]} \\
\bottomrule
\end{tabularx}
\end{table}

\begin{table}[h!]
\captionof{table}{Performance Metric Distribution of the GPT-2 language model zoo. Performance metrics are computed on train and test splits of openwebtext.\looseness-1}
\label{tab:zoo_performance_gpt2}
\centering
\tiny
\setlength{\tabcolsep}{8pt}
\begin{tabularx}{1.0\linewidth}{cccc}
\toprule
{\textbf{Train Loss}}        & {\textbf{Test Loss}}         & {\textbf{Train Perplexity}}         & {\textbf{Test Perplexity}}          \\
{$\mu \pm \sigma$ [min,max]} & {$\mu \pm \sigma$ [min,max]} & {$\mu \pm \sigma$ [min,max]} & {$\mu \pm \sigma$ [min,max]}  \\
\cmidrule(lr){1-1} \cmidrule(lr){2-2} \cmidrule(lr){3-3} \cmidrule(lr){4-4} 
0.91 $\pm$ 1.43 [0.08,10.05] & 5.97 $\pm$ 2.31 [3.86,11.29]
 & 152.29 $\pm$ 1765.46 [1.09,23065.60] & 9121.83 $\pm$ 21080.70 [54.49,79800.74]
 \\
\bottomrule
\end{tabularx}
\end{table}

\begin{table}[h!]
\captionof{table}{Performance Metric Distribution of the SciML model zoo.\looseness-1}
\label{tab:zoo_performance_sciml}
\centering
\tiny
\setlength{\tabcolsep}{48pt}
\begin{tabularx}{0.82\linewidth}{cc}
\toprule
{\textbf{Train Loss}}        &  {\textbf{Test Error}}    \\
{$\mu \pm \sigma$ [min,max]} & {$\mu \pm \sigma$ [min,max]} \\
\cmidrule(lr){1-1} \cmidrule(lr){2-2}
0.31 $\pm$ 0.18 [0.00048, 0.81] & 0.81 $\pm$ 0.34 [0.021, 1.28]  \\
\bottomrule
\end{tabularx}
\end{table}

\begin{figure}
    \centering
    \includegraphics[width=\textwidth]{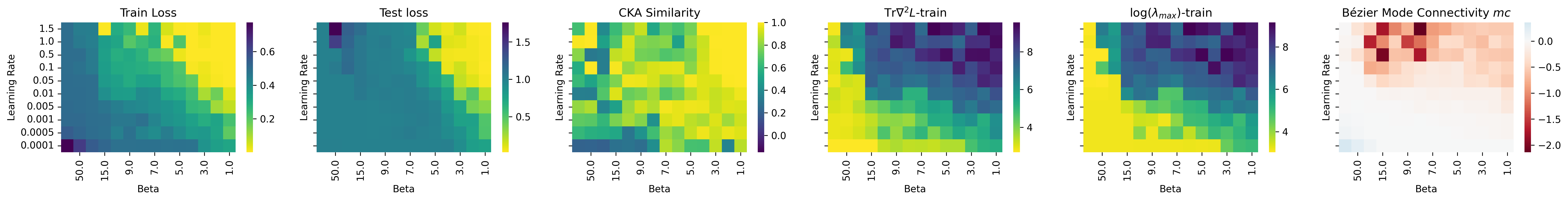}
    \caption{Phase plots for the 1D Convection PINN model zoo, showing distinct phase transitions in performance and loss-landscape metrics.}
    \label{fig:phase_plot_sciml}
\end{figure}


\clearpage

\subsection{Intended uses}

The dataset is a repository of trained deep-learning models with phase transitions. It is mainly intended to study phase transitions on populations of neural network models. For every model, we include multiple checkpoints, representing different training epochs, to allow for the study of the training procedure. We also provide loss landscape metrics, to allow researchers to relate their findings with the structure of the loss landscape. 
The dataset is intended to allow researchers to \textbf{(i)} identify phases in different model properties or applications like the weight averaging examples in the main paper; and \textbf{(ii)} evaluate existing methods that rely on pre-trained models systematically on models of different phases, to get a better understanding under which conditions methods can be expected to perform well.
Further examples of applications of our dataset are presented in our publication: model training, model property prediction, model generation, model combination, etc. 

Please note that this dataset is intended for research on populations of models, not to further improve performance on specific tasks directly. The models in our zoo were selected for their diversity in phases, not optimized for performance on their specific datasets; there may exist generating factors combinations achieving better performance with similar architectures.

The dataset is entirely synthetic and does not contain personally identifiable information or offensive content. Authors bear all responsibility in case of violation of rights.

\subsection{Hosting, Licensing, and Maintenance Plan} 
The dataset is made publicly available and licensed under the Creative Commons Attribution 4.0 International license (CC-BY 4.0). It can be accessed at \url{https://phasetransitions.modelzoos.cc}, together with the corresponding code and instructions on how to use it.


The dataset is hosted by the University of St. Gallen and will be made available to the general public with detailed instructions and examples through the project's public GitHub repository. The University of St. Gallen will ensure maintenance and long-term availability. We further plan on extending the dataset towards more architectures, tasks, and domains, and invite the community to engage. 

\end{document}